\documentclass{iosart2x}
\usepackage[ruled,lined]{algorithm2e}% provides Algorithm environment
\usepackage{graphicx}% allows for inclusion of graphic files (figures)
\usepackage[caption=false]{subfig}
\usepackage{booktabs}
\usepackage{color}
\usepackage{epstopdf}
\usepackage{xcolor}
\usepackage{amssymb}
\usepackage{amsmath}
\usepackage{capt-of}

\usepackage{tikz}
\usetikzlibrary{arrows,automata,backgrounds,decorations,fit,petri,positioning,petri,shapes,calc,patterns}
\tikzset{small/.style={level distance=10pt,sibling distance=-2pt,outer ysep=-2pt, label distance=-8pt}}
\tikzset{smaller/.style={level distance=20pt}}
\tikzset{smallish/.style={level distance=20pt,sibling distance=-2pt,outer ysep=-2pt, label distance=-8pt}}
\tikzset{node distance=5mm} % TikZ grid width and height (does not influence trees, only Petri nets and such)
\tikzset{place/.append style={circle,draw=black,thick,inner sep=0pt,minimum size=4mm,label position=below}} %TODO set label position below, emph and smaller font
\tikzset{transition/.append style={rectangle,draw=black,thick,inner sep=0pt,minimum size=4.8mm}}
\tikzset{tau/.style={transition,fill=black}}  %TAU nodes
\newcommand{\Lan}	{\mathfrak{L}}
\newcommand{\N}     {\mathbb{N}}
\newcommand{\E}     {\mathcal{E}}
\newcommand{\pre}[1]{\bullet #1}

\newlength{\hatchspread}
\newlength{\hatchthickness}
\newlength{\hatchshift}
\newcommand{\hatchcolor}{}
\tikzset{hatchspread/.code={\setlength{\hatchspread}{#1}},
	hatchthickness/.code={\setlength{\hatchthickness}{#1}},
	hatchshift/.code={\setlength{\hatchshift}{#1}},% must be >= 0
	hatchcolor/.code={\renewcommand{\hatchcolor}{#1}}}
\tikzset{hatchspread=3pt,
	hatchthickness=0.4pt,
	hatchshift=0pt,% must be >= 0
	hatchcolor=black}
\pgfdeclarepatternformonly[\hatchspread,\hatchthickness,\hatchshift,\hatchcolor]% variables
{custom north west lines}% name
{\pgfqpoint{\dimexpr-2\hatchthickness}{\dimexpr-2\hatchthickness}}% lower left corner
{\pgfqpoint{\dimexpr\hatchspread+2\hatchthickness}{\dimexpr\hatchspread+2\hatchthickness}}% upper right corner
{\pgfqpoint{\dimexpr\hatchspread}{\dimexpr\hatchspread}}% tile size
{% shape description
	\pgfsetlinewidth{\hatchthickness}
	\pgfpathmoveto{\pgfqpoint{0pt}{\dimexpr\hatchspread+\hatchshift}}
	\pgfpathlineto{\pgfqpoint{\dimexpr\hatchspread+0.15pt+\hatchshift}{-0.15pt}}
	\ifdim \hatchshift > 0pt
	\pgfpathmoveto{\pgfqpoint{0pt}{\hatchshift}}
	\pgfpathlineto{\pgfqpoint{\dimexpr0.15pt+\hatchshift}{-0.15pt}}
	\fi
	\pgfsetstrokecolor{\hatchcolor}
	\pgfusepath{stroke}
}
\pubyear{0000}
\volume{0}
\firstpage{1}
\lastpage{1}

\begin{document}

\begin{frontmatter}

%\pretitle{}
\title{Generating Time-Based Label Refinements to Discover More Precise Process Models}
\runningtitle{Generating Time-Based Label Refinements to Discover More Precise Process Models}
%\subtitle{}

% For one author:
%\author{\inits{N.}\fnms{Name1} \snm{Surname1}\ead[label=e1]{first@somewhere.com}}
%\address{Department first, \institution{University or Company name},
%Abbreviate US states, \cny{Country}\printead[presep={\\}]{e1}}
%\runningauthor{N. Surname1}

% Two or more authors:
\author[A]{\inits{N.}\fnms{Niek} \snm{Tax}\ead[label=e1]{n.tax@tue.nl}%
\thanks{Corresponding author. \printead{e1}.}},
\author[B]{\inits{E.}\fnms{Emin} \snm{Alasgarov}\ead[label=e2]{ealasgarov@bol.com}},
\author[A]{\inits{N.}\fnms{Natalia} \snm{Sidorova}\ead[label=e3]{n.sidorova@tue.nl}},
\author[C]{\inits{R.}\fnms{Reinder} \snm{Haakma}\ead[label=e4]{r.haakma@philips.com}},
and
\author[A]{\inits{W.M.P.}\fnms{Wil M.P.} \snm{van der Aalst}\ead[label=e5]{w.m.p.v.d.aalst@tue.nl}}
\runningauthor{N. Tax et al.}
\address[A]{\institution{Eindhoven University of Technology},
Eindhoven, \cny{The Netherlands}\printead[presep={\\}]{e1,e3,e5}}
\address[B]{\institution{Bol.com},
Utrecht, \cny{The Netherlands}\printead[presep={\\}]{e2}}
\address[C]{\institution{Philips Research},
	Eindhoven, \cny{The Netherlands}\printead[presep={\\}]{e4}}

\begin{abstract}
Process mining is a research field focused on the analysis of event data with the aim of extracting insights related to dynamic behavior. Applying process mining techniques on data from smart home environments has the potential to provide valuable insights into (un)healthy habits and to contribute to ambient assisted living solutions. Finding the right event labels to enable the application of process mining techniques is however far from trivial, as simply using the triggering sensor as the label for sensor events results in uninformative models that allow for too much behavior (i.e., the models are overgeneralizing). Refinements of sensor level event labels suggested by domain experts have been shown to enable discovery of more precise and insightful process models. However, there exists no automated approach to generate refinements of event labels in the context of process mining. In this paper we propose a framework for the automated generation of label refinements based on the time attribute of events, allowing us to distinguish behaviourally different instances of the same event type  based on their time attribute. We show on a case study with real-life smart home event data that using automatically generated refined labels in process discovery, we can find more specific, and therefore more insightful, process models. We observe that one label refinement could have an effect on the usefulness of other label refinements when used together. Therefore, we explore four strategies to generate useful combinations of multiple label refinements and evaluate those on three real-life smart home event logs.
\end{abstract}


\end{frontmatter}

\section{Introduction}
\label{sec:introduction}
Process mining is a fast growing discipline that combines knowledge and techniques from data mining, process modeling, and process model analysis~\cite{Aalst2016}. Process mining techniques analyze events that are logged during process execution. Today, such event logs are readily available and contain information on what was done, by whom, for whom, where, when, etc. Events can be grouped into cases (process instances), e.g., per patient for a hospital log, or per insurance claim for an insurance company. \emph{Process discovery} plays an important role in process mining, focusing on extracting interpretable models of processes from event logs. One of the attributes of the events is usually used as its label, and its values become transition/activity labels in the process models generated by process discovery algorithms.

The scope of process mining has broadened in recent years from business process management to other application domains, one of them is the analysis of events of human behavior with data originating from sensors in smart home environments \cite{Sztyler2015,Tax2016a,Tax2016b,Leotta2015,Dimaggio2016,Sztyler2016}. Table \ref{tab:example_log} shows an example of such an event log. Events in the event log are generated by, e.g., motion sensors placed in the house, power sensors placed on appliances, open/close sensors placed on closets and cabinets, etc. Particularly challenging in applying process mining in this application domain is the extraction of meaningful event labels that allow for the discovery of insightful process models. Simply using the sensor that generates an event (the \emph{sensor} column in Table \ref{tab:example_log}) as event label is shown to produce non-informative process models that overgeneralize the event log and allow for too much behavior \cite{Tax2016a}. Abstracting sensor-level events into events at the level of human activity (e.g., \emph{eating}, \emph{sleeping, etc.}) using activity recognition techniques helps to discover more behaviorally constrained and more insightful process models \cite{Tax2016b,Tax2018}. However, the applicability of this approach relies on the availability of a reliable diary of human behavior at the activity level, which is often expensive or sometimes even impossible to obtain.

Existing approaches that aim at mining temporal relations from smart home environment data \cite{Huynh2008,Galushka2006,Ogale2007,Peterson1981,Nazerfard2011,Jakkula2007} do not support the rich set of temporal ordering relations that are found in the process models \cite{Aalst2003}, which amongst others include sequential ordering, (exclusive) choice, parallel execution, and loops.

In our earlier work \cite{Tax2016a}, we showed that better process models can be discovered by taking the name of the sensor that generated the event as a starting point for the event label and then refining these labels using information on the time within the day at which the event occurred. The refinements used in \cite{Tax2016a} were based on domain knowledge, and not identified automatically from the data. In this paper, we aim at the automatic generation of semantically interpretable label refinements that can be explained to the user, by basing label refinements on data attributes of events. We explore methods to bring parts of the timestamp information to the event label in an intelligent and fully automated way, with the end goal of discovering behaviorally more precise and therefore more insightful process models. Initial work on generating label refinements based on timestamp information was started in \cite{tax2016d}. Here, we extend the work started in \cite{tax2016d} in two ways. First, we propose strategies to select a set of time-based label refinements from candidate time-based label refinements. Furthermore, add an evaluation of the technique in the form of a case study on a real-life smart home dataset.

We start by introducing basic concepts and notations used in this paper in Section \ref{sec:preliminaries}. In Section \ref{sec:approach}, we introduce a framework for the generation of event labels refinements based on the time attribute. In Section \ref{sec:case_study}, we apply this framework on a real-life smart home dataset and show the effect of the refined event labels on process discovery. In Section \ref{sec:ordering}, we address the case of applying multiple label refinements together. We continue by describing related work in Section \ref{sec:related_work} and conclude in Section \ref{sec:conclusion}.

\begin{table*}[t]
	\centering
	\caption{An example of an event log from a smart home environment.}
	\begin{tabular}{ccccc}
		\toprule
		Id & Timestamp & Address & Sensor & Sensor value\\
		\midrule
		1 & \textcolor{gray}{03/11/2015} 04:59:54 & \textcolor{gray}{Mountain Rd. 7} & Motion sensor - Bedroom  &  1\\
		2 & \textcolor{gray}{03/11/2015} 06:04:36 & \textcolor{gray}{Mountain Rd. 7} & Motion sensor - Bedroom & 1\\
		3 & \textcolor{gray}{03/11/2015} 08:45:12 & \textcolor{gray}{Mountain Rd. 7} & Motion sensor - Living room &  1\\
		4 & \textcolor{gray}{03/11/2015} 09:10:10 & \textcolor{gray}{Mountain Rd. 7} & Motion sensor - Kitchen & 1\\
		5 & \textcolor{gray}{03/11/2015} 09:12:01 & \textcolor{gray}{Mountain Rd. 7} & Power sensor - Water cooker & 1200\\
		6 & \textcolor{gray}{03/11/2015} 09:15:45 & \textcolor{gray}{Mountain Rd. 7} & Power sensor - Water cooker & 0\\
		\dots & \textcolor{gray}{03/11/2015} \dots & \textcolor{gray}{Mountain Rd. 7} & \dots & \dots \\
		\midrule
		7 & \textcolor{gray}{03/12/2015} 01:01:23 & \textcolor{gray}{Mountain Rd. 7} & Motion sensor - Bedroom &  1\\
		8 & \textcolor{gray}{03/12/2015} 03:13:14 & \textcolor{gray}{Mountain Rd. 7} & Motion sensor - Bedroom &  1\\
		9 & \textcolor{gray}{03/12/2015} 07:24:57 & \textcolor{gray}{Mountain Rd. 7} & Motion sensor - Bedroom &  1\\
		10 & \textcolor{gray}{03/12/2015} 08:34:02 & \textcolor{gray}{Mountain Rd. 7} & Motion sensor - Bedroom &  1\\
		11 & \textcolor{gray}{03/12/2015} 09:12:00 & \textcolor{gray}{Mountain Rd. 7} & Motion sensor - Living room & 1\\
		\dots & \textcolor{gray}{03/12/2015} \dots & \textcolor{gray}{Mountain Rd. 7} & \dots & \dots \\
		%\midrule
		%12 & \textcolor{gray}{03/14/2015} 03:41:46 & \textcolor{gray}{Mountain Rd. 7} & Motion sensor - Bedroom & 1\\
		%13 & \textcolor{gray}{03/14/2015} 05:00:17 & \textcolor{gray}{Mountain Rd. 7} & Motion sensor - Bedroom & 1\\
		%14 & \textcolor{gray}{03/14/2015} 08:52:32 & \textcolor{gray}{Mountain Rd. 7} & Motion sensor - Bedroom & 1\\
		%15 & \textcolor{gray}{03/14/2015} 09:30:54 & \textcolor{gray}{Mountain Rd. 7} & Motion sensor - Living room & 1\\
		%16 & \textcolor{gray}{03/14/2015} 09:35:25 & \textcolor{gray}{Mountain Rd. 7} & Power sensor - TV & 160 \\
		%17 & \textcolor{gray}{03/14/2015} 10:27:37 & \textcolor{gray}{Mountain Rd. 7} & Power sensor - TV & 0 \\
		%\dots & \textcolor{gray}{03/14/2015} \dots & \textcolor{gray}{Mountain Rd. 7} & \dots & \dots \\
		\midrule
		\dots&\dots&\dots&\dots&\dots\\
		\bottomrule
	\end{tabular}
	\label{tab:example_log}
\end{table*}

\section{Background}
\label{sec:preliminaries}
In this section, we introduce basic notions related to event logs and relabeling functions for traces and then define the notions of refinements and abstractions. We also introduce some Petri net basics.

%Logs generated by LifeLogging systems consist of sequences of activities.	
We use the usual sequence definition, and denote a sequence by listing its elements, e.g., we write $\langle a_1,a_2,\dots,a_{n} \rangle$ for a (finite) sequence $s:\{1,\dots,n\}\to A$ of elements from some alphabet $A$, where $s(i)=a_i$ for any $i \in \{1,\dots,n\}$; $|s|$ denotes the length of sequence $s$; $s_1 s_2$ denotes the concatenation of sequences $s_1$ and $s_2$. A \emph{language} $\Lan$ over an alphabet $A$ is a set of sequences over $A$.% $\Lan^p$ is the prefix closure of a language $\Lan$.

An event is the most elementary element of an event log. Let $\mathcal{I}$ be a set of event identifiers, $\mathcal{T}$ be the time domain, and $\mathcal{A}_1 \times \dots \times \mathcal{A}_{n}$ be an attribute domain consisting of $n$ attributes (e.g., resource, activity name, cost, etc.). An event is a tuple $e=(i,a_t,a_1,\dots,a_{n})$, with $i\in\mathcal{I}$, $a_t\in\mathcal{T}$, and $(a_1, \dots, a_{n})\in \mathcal{A}_1 \times \dots \times \mathcal{A}_{n}$. The \emph{event label} of an event is the attribute set $(a_1,\dots,a_n)$. Functions $\mathit{id}(e)$, $\mathit{label}(e)$, and $\mathit{time}(e)$ respectively return the id, the event label and the timestamp of event $e$. $\mathcal{E}=\mathcal{I}\times\mathcal{T}\times\mathcal{A}_1 \times \dots \times \mathcal{A}_{n}$ is a universe of events over $\mathcal{A}_1, \dots, \mathcal{A}_{n}$.
The rows of Table \ref{tab:example_log} are events from an event universe over the event attributes \emph{address}, \emph{sensor}, and \emph{sensor value}.

Events are often considered in the context of other events. We call $E\subseteq\mathcal{E}$ an \emph{event set} if $E$ does not contain multiple events with the same event identifier. The events in Table \ref{tab:example_log} together form an event set. A \emph{trace} $\sigma$ is a finite sequence formed by the events from an event set $E\subseteq{\mathcal{E}}$ that respects the time ordering of events, i.e., for all $k,m\in\N$, $1\leq k < m \leq |E|$, we have: $\mathit{time}(\sigma(k))\leq \mathit{time}(\sigma(m))$. We define the \emph{universe of traces} over event universe $\mathcal{E}$, denoted $\Sigma(\mathcal{E})$, as the set of all possible traces over $\mathcal{E}$. We omit $\mathcal{E}$ in $\Sigma(\mathcal{E})$ and use the shorter notation $\Sigma$ when the event universe is clear from the context. % The \emph{activity} column shows the result of a relabeling of the log where there is only one attribute in the event universe.

Often it is useful to partition an event set into smaller sets in which events belong together according to some criterion. We might for example be interested in discovering the typical behavior within a household over the course of a day. In order to do so, we can group together events with the same \emph{address} and the same day-part of the \emph{timestamp}, as indicated by the horizontal lines in Table \ref{tab:example_log}. For each of these event sets, we can construct a trace; timestamps define the ordering of events within the trace. For events of a trace having the same timestamps, an arbitrary ordering can be chosen within a trace.
%The \emph{address} attribute can be dropped from the attributes of the events, since the address is implied by the grouping of events.  We call such a group of events a trace, and the attributes shared amongst all events in a trace a trace attribute, with domain $T_{id}$.

An \emph{event partitioning function} is a function $\mathit{ep}: \mathcal{E} \to T_{id}$ that defines the partitioning of an arbitrary set of events $E\subseteq\mathcal{E}$ from a given event universe $\mathcal{E}$ into event sets $E_1,\ldots,E_j,\ldots$ where each $E_j$ is the maximal subset of $E$ such that for any $e_1, e_2\in E_j$, $\mathit{ep}(e_1)= \mathit{ep}(e_2)$; the value of $ep$ shared by all the elements of $E_j$ defines the value of the \emph{trace attribute} $T_{id}$. Note that multidimensional trace attributes are also possible, i.e., a combination of the name of the person performing the event activity and the date of the event, so that every trace contains activities of one person during one day. The event sets obtained by applying an event partitioning can be transformed into traces (respecting the time ordering of events).

%Traces are obtained by applying an event partitioning function to an event set. %Multiple universes of traces exist, as a trace universe $\Sigma$ depends on the event universe $\mathcal{E}$ and the event partitioning $ep$ over which it is defined.\\

%The sequence of the two events in our example event set in time-reversed order, $<$(2, 03/11/2015 09:10, 'Motion sensor - Living room', 'Mountain Rd. 7', 79), (1, 03/11/2015 08:45, 'Motion sensor - Bedroom', 'Mountain Rd. 7', 77)$>$ is not a valid trace, as it does not conform to the trace requirement of non-decreasing timestamps.
%Traces obtained by partitioning an event set $E\subseteq\mathcal{E}$ with $ep$ form an \emph{event log}. An \emph{event log} $L$ is a multiset of these traces.
An event log $L$ is a finite set of traces $L \subseteq \Sigma(\mathcal{E})$ such that $\forall\sigma\in L: \forall e_1,e_2\in\sigma: \mathit{ep}(e_1)=ep(e_2)$. $A_L\subseteq\mathcal{A}_1 \times \dots \times \mathcal{A}_{n}$ denotes the \emph{alphabet of event labels} that occur in log $L$. The traces of a log are often transformed before doing further analysis: very detailed but not necessarily informative event descriptions are transformed into some \emph{coarse-grained} and \emph{interpretable} labels. For the labels of the log in Table~\ref{tab:example_log}, the sensor values could be abstracted to \textit{on} and \textit{off}, or labels can be redefined to a subset of the event attributes, e.g., leaving the sensor values out completely.

After this relabeling step, some traces of the log can become identically labeled (the event id's would still be different). The information about the number of occurrences of a sequence of labels in an event log is highly relevant for process mining, since it allows process discovery algorithms to differentiate between the mainstream behavior of a process (i.e., frequently occurring behavioral patterns) and the exceptional behavior.

%\newdef[Trace Relabeling Function]
Let $\E_1, \E_2$ be event universes. A function $l: \E_1 \to \E_2$ is an \emph{event relabeling function} when it satisfies $\mathit{id}(e)=\mathit{id}(l(e))$ and $\mathit{time}(e)=\mathit{time}(l(e))$ for all events $e\in\E_1$. A relabeling function can be used to obtain more useful event labels than the full set of event attribute values, by lifting those elements of the attribute space to the label that result in strong ordering relations in the resulting log. We lift $l$ to event logs. Let $\E,\E_1,\E_2$ be event universes with $\E,\E_1,\E_2$ being pairwise different. Let $l_1: \E \to \E_1$ and $l_2: \E \to \E_2$ be event relabeling functions. Relabeling function $l_1$ is a \emph{refinement} of relabeling function $l_2$, denoted by $l_1\preceq l_2$, iff $\forall_{e_1,e_2\in \E}:\mathit{label}(l_1(e_1))=\mathit{label}(l_1(e_2))\implies \mathit{label}(l_2(e_1))=\mathit{label}(l_2(e_2))$; $l_2$ is then called an \emph{abstraction} of $l_1$.

The goal of process discovery is to discover a process model that represents the behavior seen in an event log. The activities/transitions in this discovered process model describe allowed orderings over the labels of the events in the event logs. A frequently used process modeling notation in the process mining field is the Petri net notation~\cite{Murata1989}. Petri nets are directed bipartite graphs consisting of transitions and places, connected by arcs. Transitions represent activities, while places represent the enabling conditions of transitions. Labels are assigned to transitions to indicate the type of activity that they model. A special label $\tau$ is used to represent invisible transitions, which are only used for routing purposes and not recorded in the log.

\begin{figure}[t]
	\centering
	\resizebox{\columnwidth}{!}{
	\begin{tikzpicture}
	[node distance=0.9cm,
	on grid,>=stealth',
	bend angle=20,
	auto,
	every place/.style= {minimum size=4mm},
	every transition/.style = {minimum size = 4.5mm}
	%transitionH/.style={rectangle, thick, fill=lightgray, minimum width=3mm, inner ysep=9pt }
	]
	\node [place, tokens = 1] (p1) [label=below:$p_1$]{};
	\node [transition] (b) [label=below:$t_1$, right = of p1] {$A$}
	edge [pre] node[auto] {} (p1);
	\node [place] (p2) [label=below:$p_2$, right = of b]{}
	edge [pre] node[auto] {} (b);
	\node [transition] (c) [label=below:$t_3$, below right = of p2] {$C$}
	edge [pre] node[auto] {} (p2);
	\node [transition] (c2) [label=below:$t_2$, above right = of p2] {$B$}
	edge [pre] node[auto] {} (p2);
	\node [place] (p3) [label=below:$p_3$, above right = of c]{}
	edge [pre] node[auto] {} (c)
	edge [pre] node[auto] {} (c2);
	\node [transition] (d) [label=below:$t_4$, right = of p3] {$D$}
	edge [pre] node[auto] {} (p3);
	\node [place] (p4) [label=above:$p_4$, below right = of d]{}
	edge [pre] node[auto] {} (d);
	\node [transition] (e) [label=above:$t_5$, right = of p4] {$F$}
	edge [pre] node[auto] {} (p4);
	\node [place] (p6) [label=above:$p_6$, right = of e]{}
	edge [pre] node[auto] {} (e);
	\node [place] (p5) [label=above:$p_5$, above right = of d]{}
	edge [pre] node[auto] {} (d);
	\node [transition] (f) [label=above:$t_6$, right = of p5] {$E$}
	edge [pre] node[auto] {} (p5);
	\node [place] (p7) [label=above:$p_7$, right = of f]{}
	edge [pre] node[auto] {} (f);
	\node [transition] (g) [label=below:$t_7$, minimum width=3mm, fill=lightgray, below right = of p7] {}
	edge [pre] node[auto] {} (p7)
	edge [pre] node[auto] {} (p6);
	\node [place,pattern=custom north west lines,hatchspread=1.5pt,hatchthickness=0.25pt,hatchcolor=gray] (p8) [label=below:$p_8$, right = of g]{}
	edge [pre] node[auto] {} (g);
	
	\end{tikzpicture}}
	\caption{An example Petri net.}
	\label{fig:example_petri_net}
\end{figure}
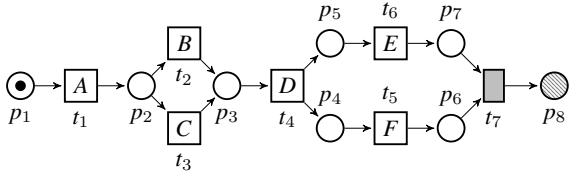

A \emph{labeled Petri net} $N=\langle P,T,F,A_M,\ell\rangle$ is a tuple where $P$ is a finite set of places, $T$ is a finite set of transitions such that $P \cap T = \emptyset$, $F \subseteq (P \times T) \cup (T \times P)$ is a set of directed arcs, called the flow relation, $A_M$ is an alphabet of labels representing activities, with $\tau \notin A_M$ being a label representing invisible events, and $\ell:T\rightarrow A_M\cup \{\tau\}$ is a labeling function that assigns a label to each transition. For a node $n \in P \cup T$ we use $\bullet n$ and $n \bullet$ to denote the set of input and output nodes of $n$, defined as $\bullet n =\{n'\mid(n',n)\in F\}$ and $n \bullet =\{n'\mid(n,n')\in F\}$. An example of a Petri net can be seen in Figure~\ref{fig:example_petri_net}, where circles represent places and rectangles represent transitions. Gray transitions having a smaller width represent invisible, or $\tau$, transitions.

A state of a Petri net is defined by its \emph{marking} $M \in \mathbb{N}^{P}$ being a multiset of places. A marking is graphically denoted by putting $M(p)$ tokens on each place $p\in P$. A pair $(N,M)$ is called a \emph{marked Petri net}. State changes occur through transition firings. A transition $t$ is enabled (can fire) in a given marking $M$ if each input place $p\in \pre{t}$ contains at least one token. Once a transition fires, one token is removed from each input place of $t$ and one token is added to each output place of $t$, leading to a new marking. An \emph{accepting Petri net} is a 3-tuple $(N,M_i,M_f)$ with $N$ a labeled Petri net, $M_i$ an initial marking, and $M_f$ a final marking. Visually, places that belong to the initial marking contain a token (e.g., $p_1$ in Figure \ref{fig:example_petri_net}), and places that belong to the final marking are depicted as $\begin{tikzpicture}
[node distance=1.4cm,
on grid,>=stealth',
bend angle=20,
auto,
every place/.style= {minimum size=0.1mm},
]
\node [place,pattern=custom north west lines,hatchspread=1.5pt,hatchthickness=0.25pt,hatchcolor=gray] {};
\end{tikzpicture}$.
Many process modeling notations, including accepting Petri nets, have formal executional semantics and a model defines a \emph{language of accepting traces} $\Lan$. The language of a Petri net consists of all sequences of activities that have a firing sequence through the Petri net that starts in the initial marking and ends in the final marking. For the Petri net in Figure \ref{fig:example_petri_net}, the language of accepting traces is $\{\langle A,B,D,E,F\rangle,\langle A,B,D,F,E\rangle,\langle A,C,D,E,F\rangle,\\\langle A,C,D,F,E\rangle\}$. In words: the process starts with activity A, followed by a choice between activity B and C, followed by activity D, finally followed by activity E and F in parallel (i.e., they can occur in any order). We refer the reader to \cite{Murata1989} for a more thorough introduction of Petri nets.

For an event log $L$ and a process model $M$ we say that $L$ is \emph{fitting} on process model $M$ if $L{\subseteq}\Lan(M)$. \emph{Precision} is related to the behavior that is allowed by a process model $M$ that was not observed in the event log $L$, i.e., $\Lan(M){\setminus}L$. The aim of process discovery is to discover a process model based on and event log $L$ that has both high \emph{fitness} (i.e., it allows for the behavior seen in the log) and high \emph{precision} (i.e., it does not allow for too much behavior that was not seen in the log). Many process discovery algorithms have been proposed throughout the years, including techniques based on Integer Linear Programming and the theory of regions \cite{Werf2008}, Inductive Logic Programming \cite{Goedertier2009}, maximal pattern mining \cite{Liesaputra2015}, or based on heuristic techniques \cite{Weijters2011,Augusto2016}. We refer the reader to \cite{Aalst2016} for a thorough introduction of several process discovery techniques.

In process discovery tasks on event logs from the business process management domain, events are often simply relabeled to the value of an \emph{activity name} attribute, which stores a generally understood name for the event (e.g., \emph{receive loan application}, or \emph{decide on building permit application}). However, event logs from the smart home environment domain generally do not contain a single attribute such that relabeling on that attribute enables the discovery of insightful process models \cite{Tax2016a}. In this paper we explore a strategy to refine an event label that is based on the name of the sensor in a smart home with information about the time in the day at which the sensor was triggered.

\section{A Framework for Time-based Label Refinements}
\label{sec:approach}
In this section, we describe a framework to generate an event label that contains partial information about the event timestamp, in order to make the event labels more specific while preserving interpretability. Note that by bringing time-in-the-day information to the event label we aim at uncovering daily routines of the person under study. We take a clustering-based approach by identifying dense areas in time-space for each label. The time part of the timestamps consists of values between $\textit{00:00:00}$ and $\textit{23:59:59}$, equivalent to the timestamp attribute from Table \ref{tab:example_log} with the day-part of the timestamp removed. This timestamp can be transformed into a real number time representation in the interval $[0,24)$. We chose to apply soft clustering (also referred to as fuzzy clustering), which has the benefit of assigning to each data point a likelihood of belonging to each cluster. A well-known approach to soft clustering is based on the combination of the Expectation-Maximization (EM)~\cite{Dempster1977} algorithm with mixture models, which are probability distributions consisting of multiple components of the same probability distribution. Each component in the mixture represents one cluster, and the probability of a data point belonging to that cluster is the probability that this cluster generated that data point. The EM algorithm is used to obtain a maximum likelihood estimate of the mixture model parameters, i.e., the parameters of the probability distributions in the mixture.

\begin{figure}[t]
	\centering
	\includegraphics[width=\columnwidth]{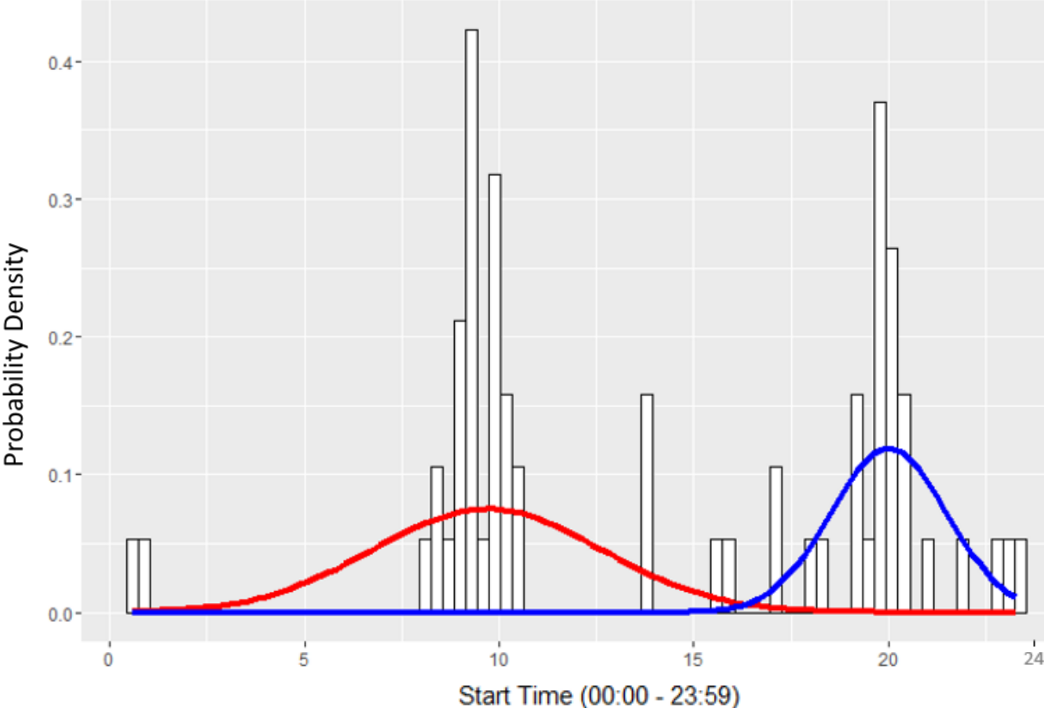}
	\caption{The histogram representation and a Gaussian Mixture Model fitted to timestamps values of the plates cupboard sensor in the Van Kasteren~\cite{Kasteren2008} dataset.}
	\label{fig:problem_data_set}
\end{figure}

A well-known type of mixture model is the Gaussian Mixture Model (GMM), where the components in the mixture distributions are normal distributions. The data space of time is, however, non-Euclidean: it has a circular nature, e.g., $23.99$ is closer to $0$ than to $23$. This circular nature of the data space introduces problems for GMMs. Figure~\ref{fig:problem_data_set} illustrates the problem of GMMs in combination with circular data by plotting the timestamps of the bedroom sensor events of the Van Kasteren \cite{Kasteren2008} real-life smart home event log. The GMM fitted to the timestamps of the sensor events consists of two components, one with the mean at $9.05$ (in red) and one with a mean at $20$ (in blue). The histogram representation of the same data shows that some events occurred just after midnight, which on the clock is closer to $20$ than to $9.05$. The GMM, however, is unaware of the circularity of the clock, which results in a mixture model that seems inappropriate when visually comparing it with the histogram. The standard deviation of the mixture component with a mean at $9.05$ is much higher than one would expect based on the histogram as a result of the mixture model trying to explain the data points that occurred just after midnight. The field of circular statistics (also referred to as directional statistics), concerns the analysis of such circular data spaces (cf. \cite{Mardia2009}). In this paper, we use a mixture of \emph{von Mises} distributions to capture the daily patterns.

Here, we introduce a framework for generating refinements of event labels based on time attributes using techniques from the field of circular statistics. This framework consists of three stages to apply to the set of timestamps of a sensor:
\begin{description}
	\setlength\itemsep{0em}
	\item[Data-model pre-fitting stage]{A known problem with many clustering techniques is that they return clusters even when the data should not be clustered. In this stage, we assess if the events of a certain sensor should be clustered at all, and if so, how many clusters it contains. For sensor types that are assessed to not be clusterable (i.e., the data consists of one cluster), the procedure ends and the succeeding two stages are not executed.}
	\item[Data-model fitting stage]{In this stage, we cluster the events of a sensor type by timestamp using a mixture consisting of components that take into account the circularity of the data. The clustering result obtained in the fitting stage is now a candidate label refinement. The label can be refined based on the clustering result by adding the assigned cluster to the label of the event, e.g., \emph{open/close fridge} can be relabeled into three distinct labels \emph{open/close fridge 1}, \emph{open/close fridge 2}, and \emph{open/close fridge 3} in case the timestamps of the fridge where clustered into three clusters.}
	\item[Data-model post-fitting stage]{In this stage, the quality of the candidate label refinements is assessed from both a cluster quality perspective and a process model (event ordering statistics) perspective. The label is only refined when the candidate label refinement is 1) based on a clustering that has a sufficiently good fit with the data, and 2) helps to discover a more insightful process model. If the candidate label refinement does not pass one of the two tests, the label refinement candidate will not be applied (i.e., the label will remain to only consist of the sensor name).}
\end{description}

We now proceed with introducing the three stages in detail.

%\begin{figure}[t]
%	\centering
%	\includegraphics[width=0.6\linewidth]{gfx/framework}
%	\caption{Time-based label refinement framework.}
%	\label{fig:framework}
%\end{figure}

\subsection{Data-model pre-fitting stage}
This stage consists of three procedures: a test for uniformity, a test for unimodality, and a method to select the number of clusters in the data. If the timestamps of a sensor type are consider to be uniformly distributed or follow a unimodal distribution, the data is considered to not be clusterable, and the sensor type will not be refined. If the timestamps are neither uniformly distributed nor unimodal, then the procedure for the selection of number of clusters will decide on the number of clusters used for clustering.

\subsubsection{Uniformity Check}
Rao's spacing test~\cite{Rao1976} tests the uniformity of the timestamps of the events from a sensor around the circular clock. This test is based on the idea that uniform circular data is distributed evenly around the circle, and $n$ observations are separated from each other $\frac{2\pi}{n}$ radii. The null hypothesis is that the data is uniform around the circle.

Given $n$ successive observations $f_1,\ldots,f_n$, either clockwise or counterclockwise, the test statistics $U$ for Rao's Spacing Test is defined as $U = \frac{1}{2}\sum_{i = 1}^{n}\mid T_i - \lambda \mid$, where $\lambda = \frac{2\pi}{n}$, $T_i = f_{i + 1} - f_{i}$ for $1 \le i \le n - 1$ and $T_n = (2\pi - f_n) + f_1$.

\subsubsection{Unimodality Check}
Hartigan's dip test~\cite{Hartigan1985} tests the null hypothesis that the data follows a unimodal distribution on a circle. When the null hypothesis can be rejected, we know that the distribution of the data is at least bimodal. Hartigan's dip test measures the maximum difference between the empirical distribution function and the unimodal distribution function that minimizes that maximum difference.

\subsubsection{Selecting the Number of Mixture Components}
The Bayesian Information Criterion (BIC)~\cite{Schwarz1978} introduces a penalty for the number of model parameters to the evaluation of a mixture model. Adding a component to a mixture model increases the number of parameters of the mixture with the number of parameters of the distribution of the added component. The likelihood of the data given the model can only increase by adding extra components, adding the BIC penalty results in a trade-off between the number of components and the likelihood of the data given the mixture model. BIC is formally defined as $\textit{BIC} = -2 * ln(\hat{L}) + k * ln(n)$, where $\hat{L}$ is a maximized value for the data likelihood, $n$ is the sample size, and $k$ is the number of parameters to be estimated. A lower BIC value indicates a better model. We start with one component and iteratively increase the number of components from $k$ to $k+1$ as long as the decrease in BIC is larger than 10, which is shown to be an appropriate threshold in \cite{Kass1995}.

\subsection{Data-model fitting stage}
A generic approach to estimate a probability distribution from data that lies on a circle or any other type of manifold (e.g., the torus and sphere) was proposed by Cohen and Welling in \cite{Cohen2015}. However, their approach estimates the probability distribution on a manifold in a non-parametric manner, and it does not use multiple probability distribution components, making it unsuitable as a basis for clustering.

We cluster events generated by one sensor using a mixture model consisting of components of the von Mises distribution, which is the circular equivalent of the normal distribution. This technique is based on the approach of Banerjee et al. \cite{Banerjee2005} that introduces a clustering method based on a mixture of von Mises-Fisher distribution components, which is a generalization of the $2$-dimensional von Mises distribution to $n$-dimensional spheres. A probability density function for a von Mises distribution with mean direction $\mu$ and concentration parameter $\kappa$ is defined as $\textit{pdf}(\theta \mid \mu, \kappa) = \frac{1}{2\pi I_0(\kappa)}e^{\kappa\cos(\theta - \mu)}$, where mean $\mu$ and data point $\theta$ are expressed in radians on the circle, such that $0 \le \theta \le 2\pi, ~0 \le \mu \le 2\pi, ~\kappa \ge 0$.
\textit{$I_0$} represents the modified Bessel function of order 0, defined as $I_0(k) = \frac{1}{2\pi}\int_0^{2\pi} e^{\kappa\cos(\theta)}d\theta$. As $\kappa$ approaches 0, the distribution becomes uniform around the circle. As $\kappa$ increases, the distribution becomes relatively concentrated around the mean $\mu$ and the von Mises distribution starts to approximate a normal distribution. We fit a mixture model of von Mises components using the package movMF \cite{Hornik2014} provided in R, using the number of components found with the BIC procedure of the pre-fitting stage. A candidate label refinement is created based on the clustering result, where the original label based on the sensor type is refined into a new number of distinct labels, each representing one von Mises component, where each event is relabeled according to the von Mises component that has the assigns the highest likelihood to the timestamp of that event.

\subsection{Data-model post-fitting stage}
This stage consists of two procedures: a statistical test to assess how well the clustering result fits the data, and a test to assess whether the ordering relations in the log become stronger by applying the relabeling function (i.e., whether it becomes more likely to discover a precise process model with process discovery techniques).
\subsubsection{Goodness-of-fit test}
After fitting a mixture of von Mises distributions to the sensor events, we perform a goodness-of-fit test to check whether the data could have been generated from this distribution. We describe the Watson $U^2$ statistic~\cite{Watson1962}, a goodness-of-fit assessment based on hypothesis testing.
%\begin{table}[t]
%	\centering
%	\caption{Estimated parameters for a mixture of von Mises distributions, 'Frontdoor', (van Kasteren dataset) }
%	\label{tab:frontdoor_table}
%	\begin{tabular}{cccc}
%		\toprule
%		Cluster & $\mu$ (radii) & $\kappa$ & $\alpha$ \\
%		\midrule
%		Cluster 1 & 2.6 & 4.32 & 0.39\\
%		Cluster 2 & 5.1 & 2.38 & 0.61\\
%		\bottomrule
%	\end{tabular}
%\end{table}
The Watson $U^2$ statistic measures the discrepancy between the cumulative distribution function $F(\theta)$ and the empirical distribution function $F_n(\theta)$ of some sample $\theta$ drawn from some population and is defined as $U^2 = n\int_0^{2\pi} \Big[ F_n(\theta) - F(\theta) - \int_0^{2\pi} \big\{ F_n(\phi) - F(\phi) \big\} dF(\phi) \Big]^2 dF(\theta)$.

\subsubsection{Control flow test}
The clustering obtained can be used as a label refinement where we refine the original event label into a new label for each cluster. We assess the quality of this label refinement from a process perspective using the label refinement evaluation method described in \cite{Tax2016a}. This method tests whether the \emph{log statistics} that are used internally in many process discovery algorithms become significantly more deterministic by applying the label refinement. Hence, we test whether the models become more precise after time-based label refinement. An example of such a log statistic is the \emph{direct successor} statistic: $\#_{L,>}^{+}(b,c)$ is the number of occurrences of $b$ in the traces of $L$ that are directly followed by $c$, i.e., in some $\sigma \in L, i\in\{1,\dots,|\sigma|\}$ we have $label([\sigma(i)])=b$ and $label([\sigma(i+1)])=c$, likewise, $\#_{L,>}^{-}(b,c)$ is the number of occurrences of $b$ which are not directly followed by $c$. This control-flow test \cite{Tax2016a} outputs a \emph{p-value} that indicates whether such log statistics of refined activities $a_1,a_2,\dots$ of some activity $a$ change with statistical significance. When $\#_{L,>}^{+}(b,c)=\#_{L,>}^{-}(b,c)$ the entropy of $b$ being directly followed by $c$ is 1 bit, equal to a coin toss. In addition to the p-value, the test returns an \emph{information gain} value, which indicates the ratio of the decrease in the total bits of entropy in the log statistics as a result of applying the label refinement. Information gain can be used as a selection criterion for label refinements when there are multiple sensor types that can be refined according to the three steps of this framework. While the entropy of a single log statistic cannot increase by applying a label refinement, the information gain of a refinement can still be negative when it is not useful, as it increases the number activities in the log and therefore also increases the total number of log statistics.

\section{Case Study}
\label{sec:case_study}

\begin{figure*}[t]
	\centering
	\subfloat[Original event data]{
		\includegraphics[width=\linewidth]{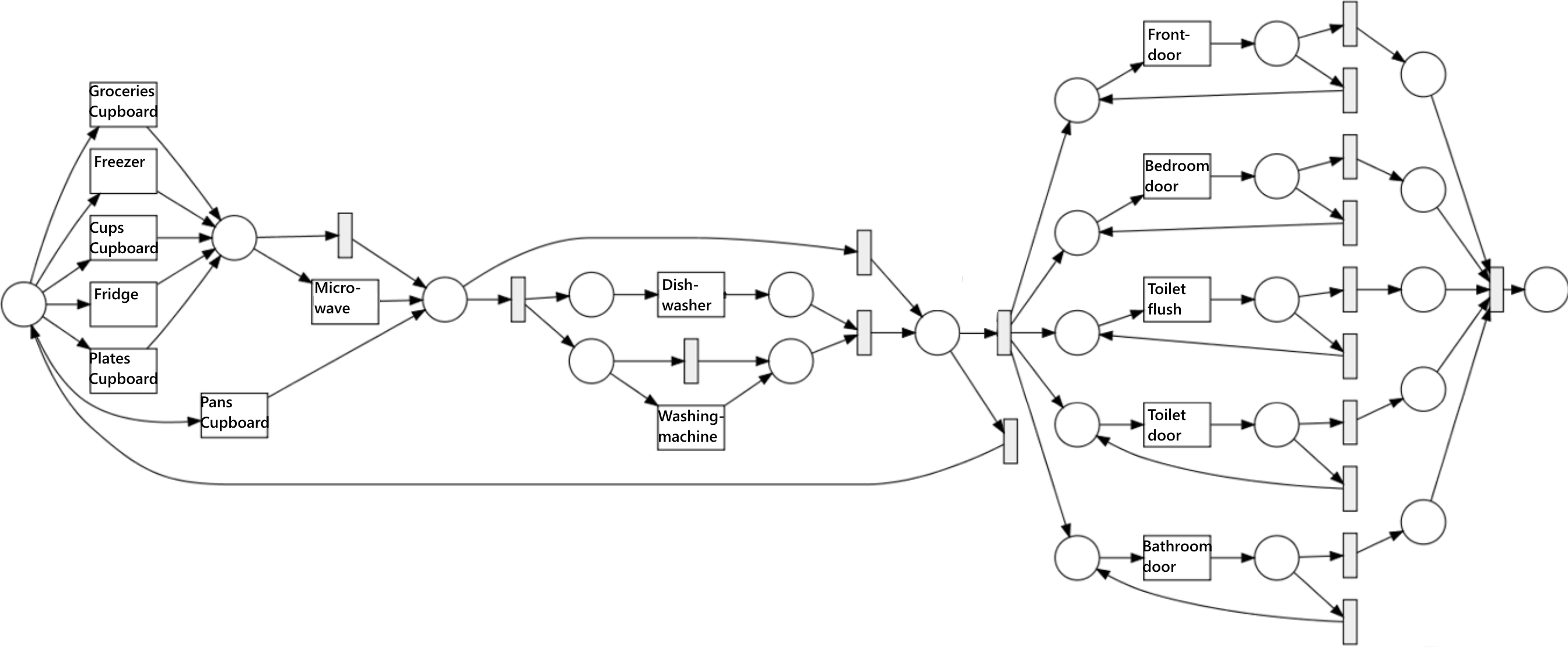}
		\label{fig:kasteren_original}
	}\\
	\subfloat[Relabeled event data]{
		\includegraphics[width=\linewidth]{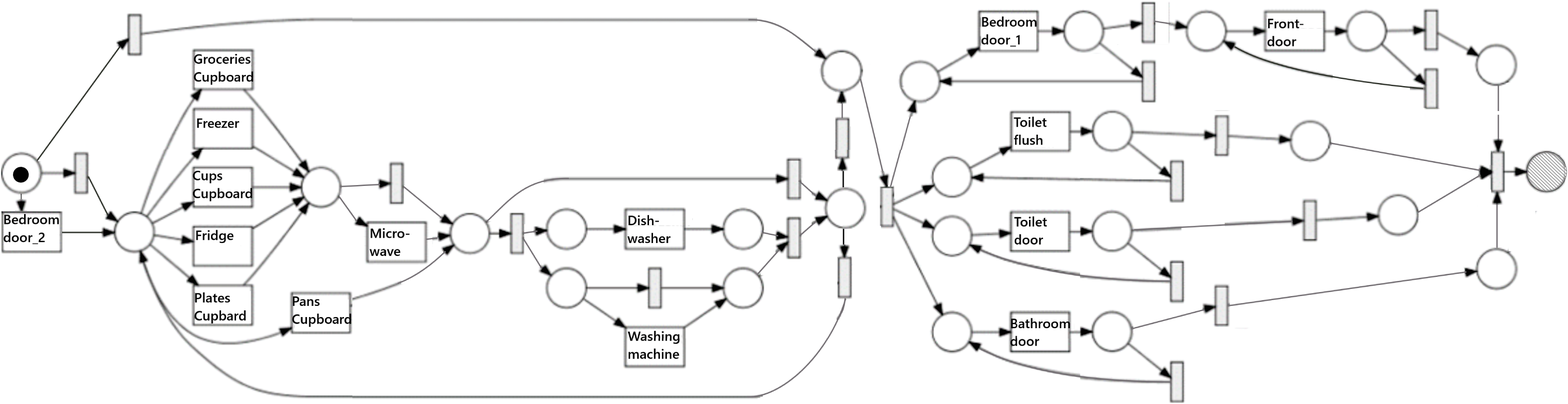}
		\label{fig:kasteren_refined}
	}
	\caption{Process models discovered on the Van Kasteren data with sensor-level labels (a) and refined labels (b) with the Inductive Miner infrequent (20\% filtering).}
\end{figure*}

\begin{figure}[t]
	\centering
	\begin{minipage}[t]{\columnwidth}
		\centering
		\includegraphics[width=\linewidth]{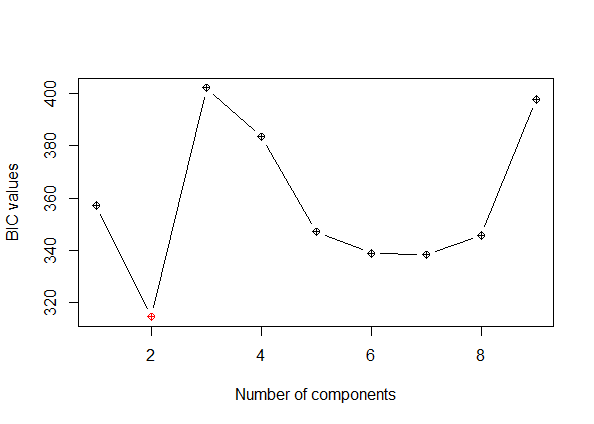}
		\caption{BIC values for different numbers of components in the mixture model.}
		\label{fig:num_component_selection_kasteren}
	\end{minipage}\hfill
	\vspace{0.2cm}
	\begin{minipage}[t]{.4\textwidth}
		\centering
		\captionof{table}{Estimated parameters for a mixture of von Mises components for bedroom door sensor events.}
		\label{tab:bedroom_components}
		\begin{tabular}{c|ccc}
			\toprule
			Cluster & $\alpha$ & $\mu$ (radii) & $\kappa$ \\
			\midrule
			Cluster 1 & 0.76 & 2.05 & 3.85\\
			Cluster 2 & 0.24 & 5.94 & 1.56\\
			\bottomrule
		\end{tabular}
	\end{minipage}
\end{figure}

We apply our time-based label refinements approach to the real-life smart home dataset described in Van Kasteren et al. \cite{Kasteren2008}. The Van Kasteren dataset consists of 1285 events divided over fourteen different sensors. We segment in days from midnight to midnight to define cases. Figure \ref{fig:kasteren_original} shows the process model discovered on this event log with the Inductive Miner infrequent \cite{Leemans2013} process discovery algorithm with 20\% filtering, which is a state-of-the-art process discovery algorithm that discovers a process model that describes the most frequent 80\% of behavior in the log. Note that this process model overgeneralizes, i.e., it allows for too much behavior. At the beginning a (possibly repeated) choice is made between five transitions. At the end of the process, the model allows any sequence over the alphabet of five activities, where each activity occurs at least once.

We illustrate the framework by applying it to the \emph{bedroom door} sensor. Rao's spacing test results in a test statistic of $241.0$ with $152.5$ being the critical value for significance level $0.01$, indicating that we can reject the null hypothesis of a uniformly distributed set of \emph{bedroom door} timestamps. Hartigan's dip test results in a p-value of $3.95\times10^{-4}$, indicating that we can reject the null hypothesis that there is only one cluster in the \emph{bedroom door} data. Figure \ref{fig:num_component_selection_kasteren} shows the BIC values for different numbers of components in the model. The figure indicates that there are two clusters in the data, as this corresponds to the lowest BIC value. Table \ref{tab:bedroom_components} shows the mean and $\kappa$ parameters of the two clusters found by optimizing the von Mises mixture model with the EM algorithm. A value of $0\equiv2\pi$ radii equals midnight. After applying the von Mises mixture model to the \emph{bedroom door} events and assigning each event to the maximum likelihood cluster we obtain a time range of [3.08-10.44] for cluster 1 and a time range of [17.06-0.88] for cluster 2. The Watson $U^2$ test results in a test statistic of $0.368$ and $0.392$ for cluster 1 and 2 respectively with a critical value of $0.141$ for a $0.01$ significance level, indicating that the data is likely to be generated by the two von Mises distributions found. The label refinement evaluation method \cite{Tax2016a} finds statistically significant differences between the events from the two \emph{bedroom door} clusters with regard to their control-flow relations with other activities in the log for 10 other activities using the significance level of $0.01$, indicating that the two clusters are different from a control-flow perspective. Figure \ref{fig:kasteren_refined} shows the process model discovered with the Inductive Miner infrequent with 20\% filtering after applying this label refinement to the Van Kasteren event log. The process model still overgeneralizes the overall process, but the label refinement does help to restrict the behavior, as it shows that the evening \emph{bedroom door} events are succeeded by one or more events of type \emph{groceries cupboard}, \emph{freezer}, \emph{cups cupboard}, \emph{fridge}, \emph{plates cupboard}, or \emph{pans cupboard}, while the morning \emph{bedroom door} events are followed by one or more \emph{frontdoor} events. It seems that this person generally goes to the bedroom in-between coming home from work and starting to cook. The loop of the \emph{frontdoor} events could be caused by the person leaving the house in the morning for work, resulting in no logged events until the person comes home again by opening the \emph{frontdoor}. Note that in Figure \ref{fig:kasteren_original} \emph{bedroom door} and \emph{frontdoor} events can occur an arbitrary number of times in any order. Figure \ref{fig:kasteren_original} furthermore does not allow for the \emph{bedroom door} to occur before the whole block of kitchen-located events at the beginning of the net. In the process mining field multiple quality criteria exist to express the fit between a process model and an event log. Two of those criteria are \emph{fitness} \cite{Rozinat2008}, which measures the degree to which the behavior that is observed in the event log can be replayed on the process model, and \emph{precision} \cite{Munoz2010}, which measures the degree to which the behavior that was never observed in the event log cannot be replayed on the process model. Low precision typically indicates an overly general process model, that allows for too much behavior. Typically we aim for process models with both high fitness and precision, therefore one can consider the harmonic mean of the two, often referred to as \emph{F-score}. The \emph{bedroom door} label refinement described above improves the precision of the process model found with the Inductive Miner infrequent (20\% filtering) \cite{Leemans2013} from $0.3577$ when applied on the original event log to $0.4447$ when applied on the refined event log and improves the F-score from $0.5245$ to $0.6156$.

\begin{figure*}
	\centering
	\includegraphics[width=\linewidth]{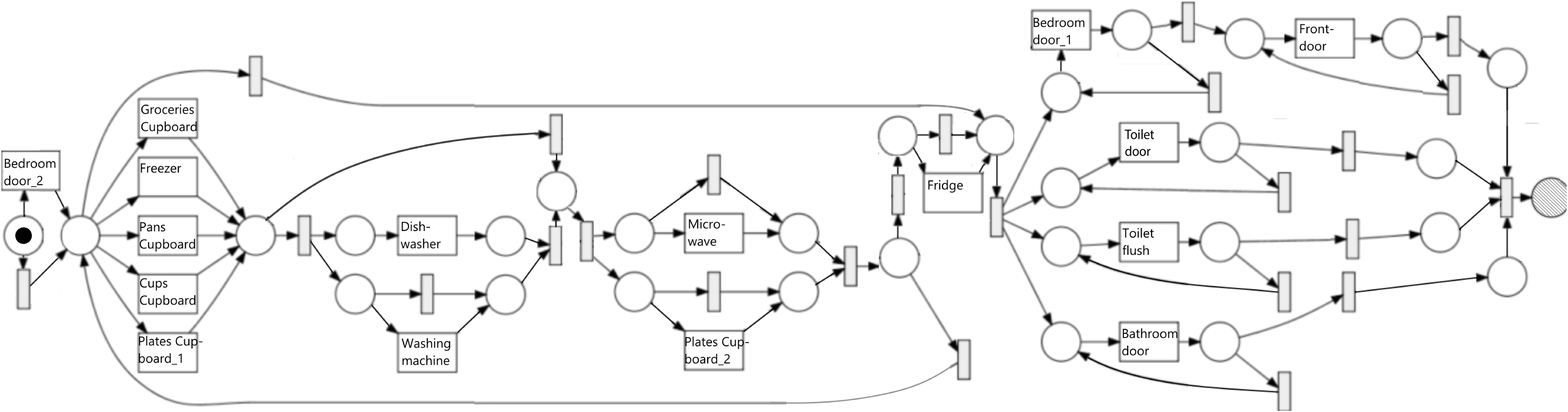}
	\caption{Inductive Miner infrequent (20\% filtering) result after a second label refinement.}
	\label{fig:kasteren_refined2}
\end{figure*}

The label refinement framework allows for refinement of multiple activities in the same log. For example, label refinements can be applied iteratively. Figure \ref{fig:kasteren_refined2} shows the effect of a second label refinement step, where \emph{Plates cupboard} using the same methodology is refined into two labels, representing time ranges [7.98-14.02] and [16.05-0.92] respectively. This refinement shows the additional insight that the evening version of the \emph{Plates cupboard} occurs directly before or after the microwave. Generating multiple label refinements, however, comes with the problem that the control-flow test \cite{Tax2016a} is sensitive to the order in which label refinements are applied. Because label refinements change the event log, it is possible that after applying some label refinement $A$, some other label refinement $B$ starts passing the control-flow test that did not pass this test before, or fails the test while it passed before. Additionally, applying one label refinement can change the information gain of applying another label refinement afterwards. For example, when $\#_{L,>}^{+}(b,c)=\#_{L,>}^{-}(b,c)$, i.e., $b$ is followed by $c$ 50\% of the time, the entropy of this log statistic is $1$, equal to a coin toss. Some label refinement $A$ which refines $b$ into $b_1,b_2$ where $b_1$ is always followed by $c$ and $b_2$ is never followed by $c$ is a good label refinement from an information gain point of view, as it decreases the entropy of the log statistic to zero. Some other label refinement $B$, which refines $c$ into $c_1,c_2$ such that all $b$'s are directly followed by $c_1$'s and never by $c_2$'s also leads to information gain. However, applying refinement $B$ after having already applied refinement $A$, does not lead to any further information gain, since refinement $A$ has already made it deterministic whether or not $b$ is followed by any $c$. Ineffective label refinements might even harm process discovery, as each refinement decreases the frequencies with which activities are observed, thereby decreasing the amount of evidence for certain control-flow relations.

\section{On the Ordering of Label Refinements}
\label{sec:ordering}
%\begin{figure}[t]
%	\centering
%	\includegraphics[width=0.8\textwidth]{gfx/experimental_resultsv3}
%	\caption{Fitness, precision, and F-score of the Inductive Miner infrequent 20\% models obtained from the original and refined versions of three event logs with stopping criterion Information Gain larger than zero.}
%	\label{fig:experimental_results2}
%\end{figure}

As shown in Section \ref{sec:case_study}, the outcome of the control flow test test of a label refinement can depend on whether other label refinements that passed the test of the pre-fitting and post-fitting stages have already been applied. Therefore, in this section, we explore the effect of the ordering of label refinements on real-life event logs. We explore this effect by evaluating four strategies to select a set of label refinements to apply to the event log. Each of the strategies assume the desired number $k$ of label refinements to be given.
\begin{description}
	\setlength\itemsep{0em}
	\item[All-at-once] In this strategy we naively ignore the influence of the interplay between label refinements on the outcome of the control flow test and select top $k$ label refinements in a single step based on their information gain that is calculated using the original event log, to which the other selected label refinements are not applied.
	\item[Greedy Search] We first apply the best label refinement in terms of information gain, then refine the event log using this label refinement, and then iterate to find the next label refinement calculating the information gain using the refined event log from the previous step.
	\item[Exhaustive Search] This strategy exhaustively tries all combinations of label refinements and searches for the label refinement combinations that jointly lead to the largest information gain. While the label refinement combinations that are found with this strategy are optimal in terms of information gain, this strategy can quickly become computationally intractable for event logs that contain many activities.
	\item[Beam Search] In Beam Search only a predetermined number $b$ (called the beam size) of best partial solutions are kept as candidates, i.e., only the best $b$ combinations in terms of information gain that were found consisting of $n$ label refinements are explored to search for a new set of $n+1$ label refinements. This is an intermediate strategy in-between greedy and exhaustive search, with greedy search being a beam search with b=1 and exhaustive search being a beam search with b=$\infty$.
\end{description}

\begin{figure*}[t]
	\centering
	\includegraphics[width=\textwidth]{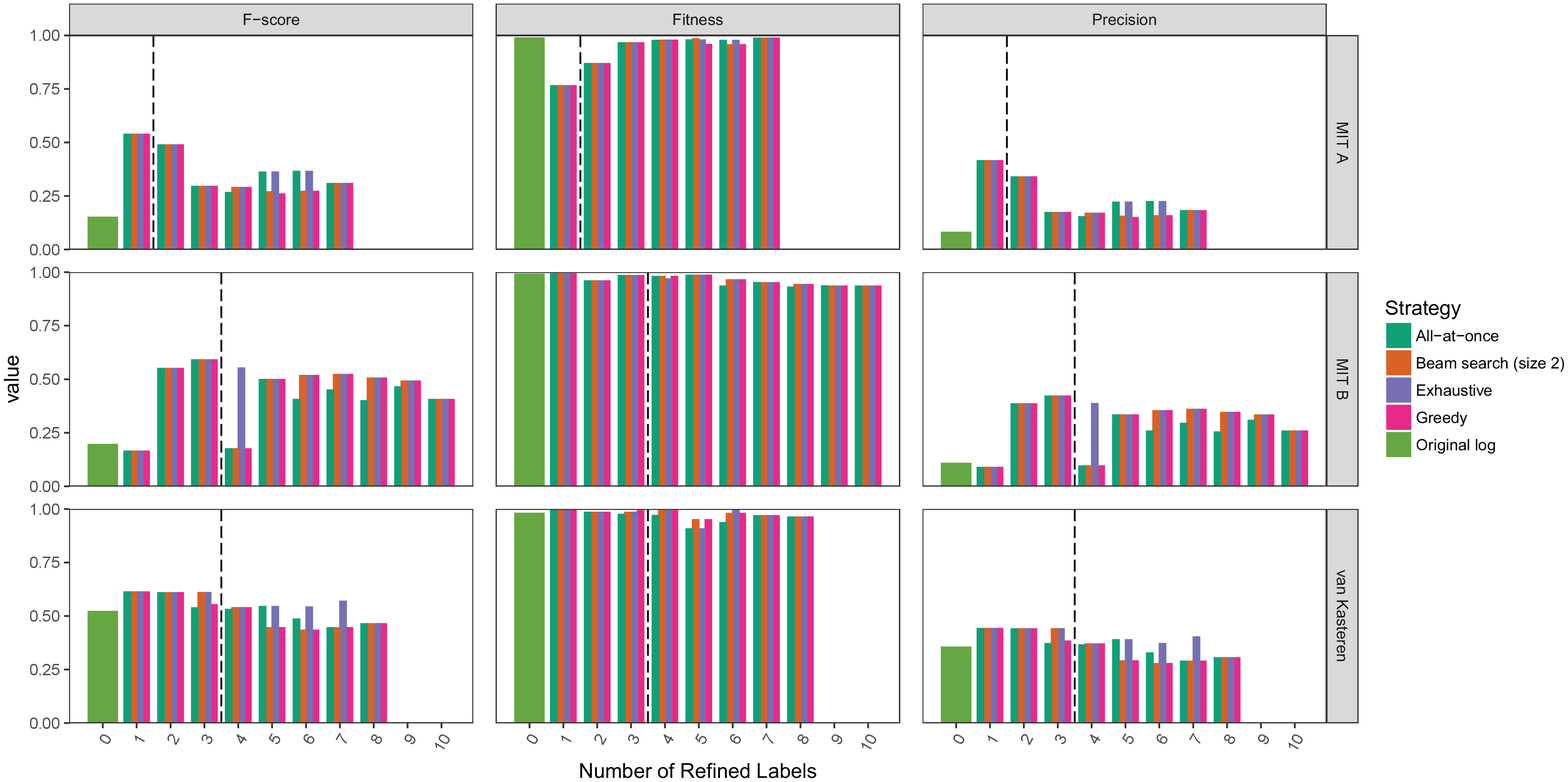}
	\caption{Fitness, precision, and F-score of the Inductive Miner infrequent (20\% filtering) models obtained from the original and refined versions of three event logs.}
	\label{fig:experimental_results}
\end{figure*}

We apply these four strategies on three event logs from the human behavior domain and measure the \emph{fitness}, \emph{precision}, and \emph{F-score} of the model discovered with the Inductive Miner infrequent \cite{Leemans2013} with 20\% filtering after each label refinement. The first event log is the Van Kasteren \cite{Kasteren2008} event log which we introduced in Section \ref{sec:case_study}. The other two event logs are two different households of a smart home experiment conducted by MIT \cite{Tapia2004}. The log \emph{Household A} of the MIT experiment contains 2701 events spread over 16 days, with 26 different sensors. The \emph{Household B} log contains 1962 events spread over 17 days and 20 different sensors. 

Figure \ref{fig:experimental_results} shows the results. On all three event logs the precision can be improved considerably through label refinements. Note that when applying only one label refinement all four strategies are identical. When refining a second label the four strategies all select the same label refinement on all three logs. Therefore the F-score, fitness, and precision for two refined labels happen to be identical. Figure \ref{fig:experimental_results} shows that for the MIT household A data set there are 7 sensor types that can be refined, i.e., they passed the statistical tests of the pre-fitting stage and their obtained clustering passed the goodness-of-fit test. For the MIT household B data set there are 10 activities that can be refined and for the are 8 activities that can be refined for the Van Kasteren data set. However, since the F-score for all strategies drops again after a few label refinements, not all of those label refinements lead to better process models. The four strategies perform very similar in terms of F-score. Exhaustive search outperforms the other strategies for a few refinements on some logs, however, such improvements come with considerable computation times. On the MIT household B log, which has 10 possible label refinements, it takes about 25 minutes on an Intel i7 processor to evaluate all possible combinations of refinements. On logs with even more possible refinements the exhaustive strategy can quickly become computationally infeasible. The all-at-once strategy, which is computationally very fast and only takes milliseconds to compute, shows almost identical performance for MIT household A and Van Kasteren. When making six or more refinements on the MIT household B log, the performance of the all-at-once strategy lags behind the other strategies, indicating that the label refinements that were applied earlier cause the later label refinements to be less effective. However, the optimum in F-score for this log lies at three refinements, therefore the sixth refinement, where a performance difference between non-exhaustive strategies emerges should not be performed with any of the strategies in the first place.

Since the F-score decreases again when applying too many label refinements it is important to have a stopping criterion that prevents refining the event log too much. The dashed line in Figure \ref{fig:experimental_results} shows the results when we only refine a label when the information gain of the refinement is larger than zero. On the MIT households A and B logs this stopping criterion causes all strategies to stop at the best combination of label refinements in F-score, consists respectively of one and three refinements. This indicates that the control flow test \cite{Tax2016a} provides a useful stopping criterion for label refinements.

All strategies except the exhaustive search strategy suggest as the fourth refinement for MIT B a refinement that decreases the F-score sharply, to increase it again with a fifth refinement. This is caused by an unhelpful refinement being found as the fourth refinement by those strategies, which causes the frequencies to drop below the filtering threshold of the Inductive Miner, leading to a model that is less precise. At the fifth refinement, the follows statistics of other activities drop as well, causing the follows statistics that dropped in the fourth refinement to be relatively higher and above the threshold again. On the Van Kasteren log the optimum in F-score is to make only one refinement, although the F-score after applying the second and third refinement as found by the exhaustive and beam search is almost identical. The all-at-once strategy stops after applying only two refinements while the other strategies apply a third refinement. The best refinement combination found with the all-at-once strategy using the stopping criterion is identical to the refinement combination found with the other strategies, suggesting that in practice the differences between the four approaches are small. On real-life smart home environment event logs the effect that one label refinement influences the control flow test outcome of others is limited.

\begin{figure*}[t]
	\centering
	\includegraphics[width=\textwidth]{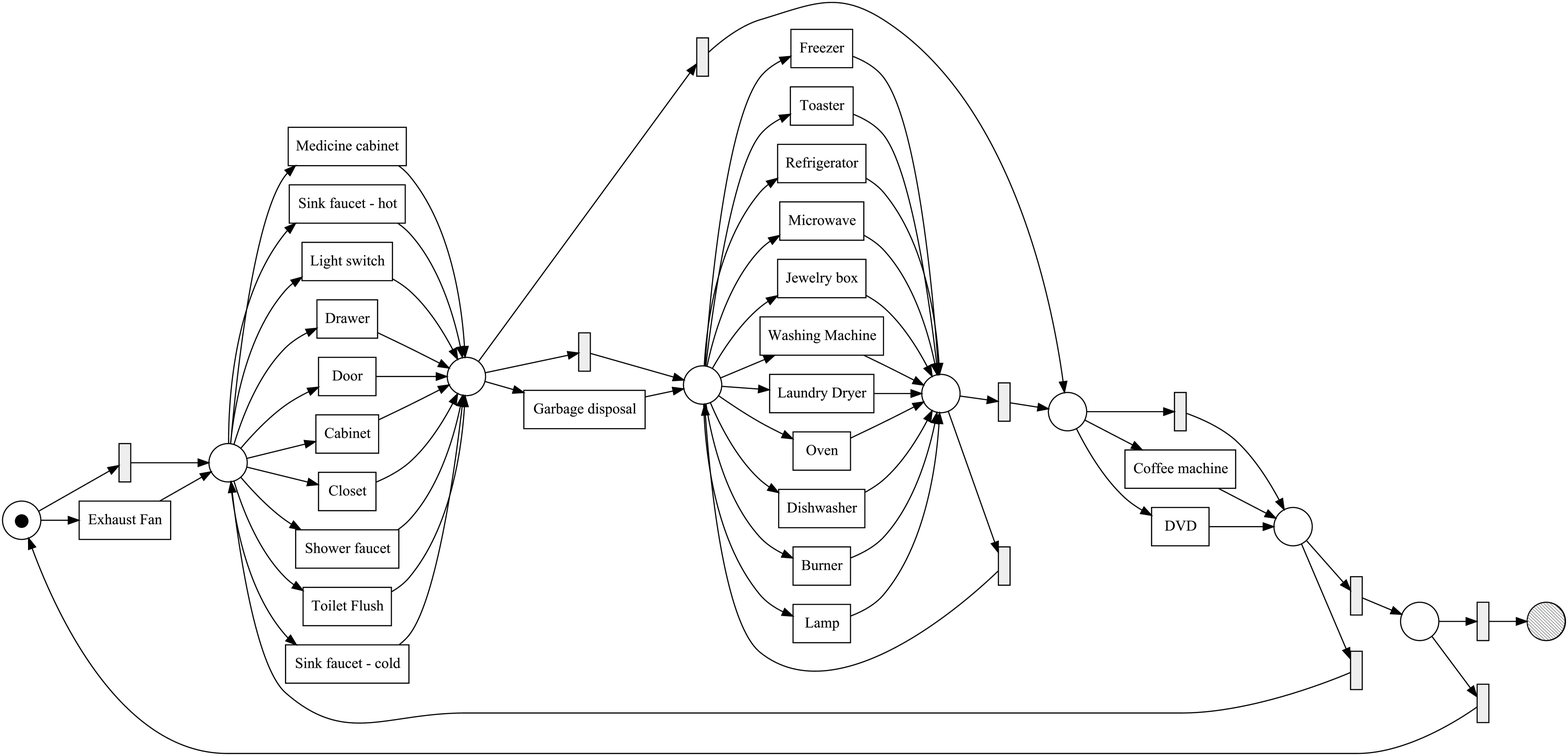}
	\caption{The Inductive Miner infrequent (20\% filtering) process model discovered from the original MIT A event log.}
	\label{fig:mit_a_original}
\end{figure*}

\begin{figure*}[t]
	\centering
	\includegraphics[width=\textwidth]{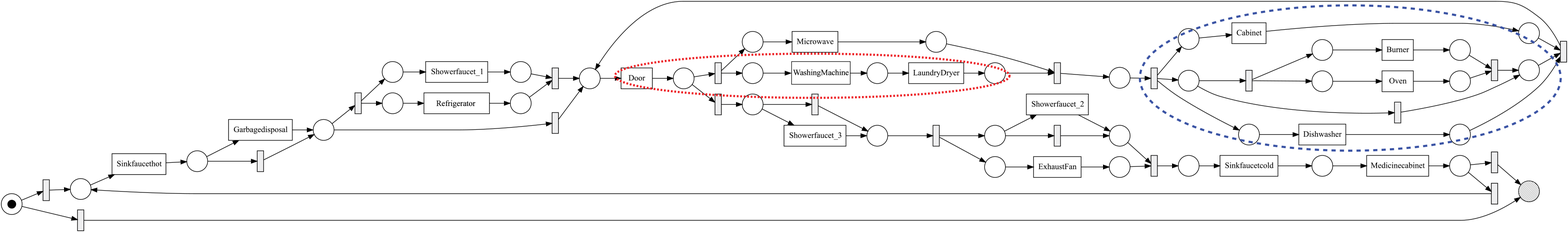}
	\caption{The Inductive Miner infrequent (20\% filtering) process model discovered from the refined MIT A event log.}
	\label{fig:mit_a_refined}
\end{figure*}

Figures \ref{fig:mit_a_original} and \ref{fig:mit_a_refined} shows the process model that are discovered with the Inductive Miner infrequent with 20\% filtering respectively from the original MIT household A event log and the event log obtained after applying the optimal combination of label refinements found in the results of Figure \ref{fig:experimental_results}. Because of the silent transitions, the process model discovered from the original event log allows for almost all orderings over the sensor types. Even though the transition labels in the process model discovered from the refined event log are not readable because of the size, it is clear from the structure of the process model that it is much more behaviorally specific, containing a mix of sequential orderings, parallel blocks, and choices over the sensor types. Especially interesting is the part indicated by the blue dashed ellipse, which contains a parallel block consisting of a \emph{cabinet}, the \emph{oven} and \emph{burner}, and the \emph{dishwasher}, showing a clearly recognizable cooking routine. Furthermore, the part indicated by the red dotted ellipse indicates a sequentially ordered part, consisting of some \emph{door} sensor registering the opening of a door, followed by starting the \emph{washing machine} and then the \emph{laundry dryer}.

The time-based label refinement generation framework as well as the four strategies to generate multiple label refinements on the same event log are implemented and publicly available in the process mining toolkit ProM \cite{Dongen2005} as part of the \emph{LabelRefinements}\footnote{\url{https://svn.win.tue.nl/repos/prom/Packages/LabelRefinements/}} package.

\section{Related Work}
\label{sec:related_work}
We classify related work into three categories. The first category of related work concerns techniques from the process mining field, specifically focusing on techniques that, like our approach, focus on refining activity labels. The second category of related work, also originating from the process mining area, focuses on the interplay between ordering between process activities and external information, such as time. The third category of related work originates from the ambient intelligence and smart home environments field, focusing on work on mining temporal relations between human activities. We use these three categories to structure this section.

\subsection{Label Splits and Refinements in Process Mining}
The task of finding refinements of event labels in the event log is closely related to the task of mining process models with duplicate activities, in which the resulting process model can contain multiple transitions/nodes with the same label. From the point of view of the behavior allowed by a process model, it makes no difference whether a process model is discovered on an event log with refined labels, or whether a process model is discovered with duplicate activities such that each transition/node of the duplicate activity precisely covers one version of the refined label. However, a refined label may also provide additional insights as the new labels are explainable in terms of time.
The first process discovery algorithm capable of discovering duplicate tasks was proposed by Herbst and Karagiannis in 2004 \cite{Herbst2004}, after which many others have been proposed, including the Evolutionary Tree Miner \cite{Buijs2012}, the $\alpha^*$-algorithm \cite{Li2007}, the $\alpha^\#$-algorithm \cite{Gu2008}, the EnhancedWFMiner \cite{Folino2009}. An alternative approach has been proposed by V\'{a}zques-Barreiros \cite{Vazquez-Barreiros2015} et al., who describe a local search based approach to repair a process model to include duplicate activities, starting from an event log and a process model without duplicate activities. Existing work on mining models with duplicate activities all base their duplicate activities on how well the event log fits the process model, and do not try to find semantic differences between the different versions of the activities in the form of attribute differences.

The work that is closest to our work is the work by Lu et al. \cite{Lu2016}, who describe an approach to pre-process an event log by refining event labels with the goal of discovering a process model with duplicate activities. The method proposed by Lu et al., however, does not base the relabelings on data attributes of those events and only uses the control flow context, leaving uncertainty whether two events relabeled differently are actually semantically different.

\subsection{Data-Aware Process Mining}
Another area of related work is data-aware process mining, where the aim is to discover rules with regard to data attributes of events that decide decision points in the process. De Leoni and van der Aalst \cite{DeLeoni2013} proposed a method that discovers data guards for decision points in the process based on alignments and decision tree learning. This approach relies on the discovery of a behaviorally well-fitting process model from the original event log. When only overgeneralizing process models (i.e., allowing for too much behavior) can be discovered from an event log, the correct decision points might not be present in the discovered process model at all, resulting in this approach not being able to discover the data dependencies that are in the event log. Our label refinements use data attributes prior to process discovery to enable the discovery of more behaviorally constrained process models by bringing parts of the event attribute space to the event label.

\subsection{Temporal Relation Mining for Smart Home Environments}
Galushka et al. \cite{Galushka2006} provide an overview of temporal data mining techniques and discuss their applicability to data from smart home environments. Many of the techniques described in the overview focus on real-valued time series data, instead of discrete sequences which we assume as input in this work. For discrete sequence data, Galushka et al. \cite{Galushka2006} propose the use of sequential rule mining techniques, which can discover rules of the form ``\emph{if event \textbf{a} occurs then event \textbf{b} occurs with time \textbf{T}''}.

Huynh et al. \cite{Huynh2008} proposed to use topic modeling to mine activity patterns from sequences of human events. Topic modeling originates from the field of natural language processing and addresses the challenge to find topics in textual documents and assign a distribution over these topics to each document. However, the discovered topics do not represent the human activities in terms of control-flow ordering constructs like sequential ordering, concurrent execution, choices, and loops. 

Ogale et al. \cite{Ogale2007} proposed an approach to describe the temporal relation between human behavior activities from video data using context-free grammars, using the human poses extracted from the video as the alphabet. Like Petri nets, context-free grammars define a formal language over its alphabet. However, Petri nets have a graphical representation, which is lacking for grammars. Furthermore, as shown by Peterson \cite{Peterson1981}, Petri net languages are a subclass of context-sensitive languages, and some Petri net languages are not context-free. This indicates that some relations over activities that can be expressed in Petri nets cannot be expressed in a context-free grammar.

One particularly related technique, called \emph{TEmporal RElation Discovery of Daily Activities (TEREDA)}, was proposed by Nazerfard et al. \cite{Nazerfard2011}. TEREDA leverages temporal association rule mining techniques to mine ordering relations between activities as well as patterns in their timestamp and duration. The ordering relations between activities that are discovered by TEREDA are restricted to the form ``\emph{activity} \textbf{a} \emph{follows activity} \textbf{b}'', where our proposed approach of modeling the relations with Petri nets allow for modeling of more complex relations between larger number of activities, such as: ``\emph{the occurrences of activity} \textbf{b} \emph{that are preceded by activity} \textbf{a} \emph{are followed by both activity} \textbf{d} \emph{and} \textbf{e}\emph{, but in arbitrary order''}. The patterns in the timestamps are obtained by fitting a Gaussian Mixture Model (GMM) with the Expectation-Maximization (EM) algorithm, thereby ignoring problems caused by the circularity of the 24-hour clock introduced in this paper.

Jukkala and Cook \cite{Jakkula2007} propose a method to mine temporal relations between activities from smart home environments logs where the temporal relation patterns are expressed in \emph{Allen's interval algebra} \cite{Allen1983}. Allen's interval algebra allows the expression of thirteen distinct types of temporal relations between two activities based on both the \emph{start} and \emph{end} timestamps of these activities. The approach of Jukkala and Cook \cite{Jakkula2007} is limited to describing the relations between pairs of activities, and more complex relations between three or higher numbers of activities cannot be discovered. The aim of mining the patterns in Allen's interval algebra representation is to increase the accuracy of activity recognition systems, while our goal is knowledge discovery.
%\begin{figure}
%	\includegraphics[width=0.8\linewidth]{gfx/gmm_tv_duration}
%	\caption{Gaussian Mixture Model with three components fitted to the durations of TV events}
%	\label{fig:gmm_tv}
%\end{figure}
Several papers from the process mining area have focused on mining temporal relations between activities from smart home event logs. Leotta et al. \cite{Leotta2015} postulate three main research challenges for the applicability of process mining technique for smart home data. One of those three challenges is to improve process mining techniques to address the less structured nature of human behavior as compared to business processes. Our technique addresses this challenge, as the time-based label refinements help in uncovering relations between activities with process mining techniques that could not be found without applying time-based label refinements.

DiMaggio et al. \cite{Dimaggio2016} and Sztyler et al. \cite{Sztyler2015,Sztyler2016} propose to mine Fuzzy Models \cite{Gunther2007} to describe the temporal relations between human activities. The Fuzzy Miner \cite{Gunther2007}, a process discovery algorithm that mines a Fuzzy Model from an event log, is a process discovery algorithm that is designed specifically for weakly structured processes. However, Fuzzy Models, in contrast to Petri nets, do not define a formal language over the activities, and are therefore not precise on what activity orderings are allowed and which are not. While mining a Fuzzy Model description of human activities is less challenging compared to mining a process model with formal semantics, it is also limited in the insights that can be obtained from it.

Finally, insights in the human routines can be obtained through the discovery of \emph{Local Process Models} \cite{Tax2016c}, which bridges process mining and sequential pattern mining by finding patterns that include high-level process model constructs such as (exclusive) choices, loops, and concurrency. However, Local Process Models, as opposed to process discovery, only give insight into frequent subroutines of behavior and do not provide the global picture of the behavior throughout the day from start to end.

\section{Conclusion \& Future Work}
\label{sec:conclusion}
We have proposed a framework based on techniques from the field of circular statistics to refine event labels automatically based on their timestamp attribute. We have shown on a real-life event log that this framework can be used to discover label refinements that allow for the discovery of more insightful and behaviorally more specific process models.
Additionally, we explored four strategies to search combinations of label refinements. We found that the difference between an all-at-once strategy, which ignores that one label refinement can have an effect on the usefulness of other label refinements, and other more computationally expensive strategies is often limited.
An interesting area of future work is to explore the use of other types of event data attributes to refine labels, e.g., power values of sensors. A next research step would be to explore label refinements based on a combination of data attributes combined. This introduces new challenges, such as the clustering on partially circular and partially Euclidean data spaces. Additionally, other time-based types of circles than the daily circle described in this paper, such as the week, month, or year circle, are worth investigating.

\nocite{*} 
% if your bibliography is in bibtex format, use those commands:
\bibliographystyle{ios1}           % Style BST file.
\bibliography{ais_template}        % Bibliography file (usually '*.bib')

\begin{thebibliography}{53}
% BibTex style file: ios1.bst, 2017-04-26
\ifx \bisbn   \undefined \def \bisbn  #1{ISBN #1}\fi
\ifx \binits  \undefined \def \binits#1{#1} \fi
\ifx \bauthor  \undefined \def \bauthor#1{#1} \fi
\ifx \bjtitle  \undefined \def \bjtitle#1{\textit{#1}}\fi
\ifx \batitle  \undefined \def \batitle#1{#1} \fi
\ifx \bctitle  \undefined \def \bctitle#1{#1} \fi
\ifx \bvolume  \undefined \def \bvolume#1{\textbf{#1}}\fi
\ifx \byear  \undefined \def \byear#1{#1} \fi
\ifx \bissue  \undefined \def \bissue#1{#1} \fi
\ifx \bfpage  \undefined \def \bfpage#1{#1} \fi
\ifx \blpage  \undefined \def \blpage #1{#1} \fi
\ifx \burl  \undefined \def \burl#1{#1} \fi
\ifx \doiurl  \undefined \def \doiurl#1{#1} \fi
\ifx \betal  \undefined \def \betal{et al.} \fi
\ifx \binstitute  \undefined \def \binstitute#1{#1} \fi
\ifx \beditor  \undefined \def \beditor#1{#1} \fi
\ifx \bpublisher  \undefined \def \bpublisher#1{#1} \fi
\ifx \bbtitle  \undefined \def \bbtitle#1{\textit{#1}} \fi
\ifx \bedition  \undefined \def \bedition#1{#1} \fi
\ifx \bseriesno  \undefined \def \bseriesno#1{#1} \fi
\ifx \blocation  \undefined \def \blocation#1{#1} \fi
\ifx \bsertitle  \undefined \def \bsertitle#1{#1} \fi
\ifx \bsnm \undefined \def \bsnm#1{#1} \fi
\ifx \bsuffix \undefined \def \bsuffix#1{#1} \fi
\ifx \bparticle \undefined \def \bparticle#1{#1} \fi
\ifx \barticle \undefined \def \barticle#1{#1} \fi
\ifx \botherref \undefined \def \botherref #1{#1} \fi
\ifx \url \undefined \def \url#1{#1} \fi
\ifx \bchapter \undefined \def \bchapter#1{#1} \fi
\ifx \bbook \undefined \def \bbook#1{#1} \fi
\ifx \bcomment \undefined \def \bcomment#1{#1} \fi
\ifx \oauthor \undefined \def \oauthor#1{#1} \fi
\ifx \citeauthoryear \undefined \def \citeauthoryear#1{#1} \fi
\ifx \texttildelow  \undefined \def \texttildelow{\symbol{126}} \fi
\def \endbibitem {}
\ifx \bconflocation  \undefined \def \bconflocation#1{#1} \fi

\bibitem{Aalst2016}
\begin{bbook}
\bauthor{\binits{W.M.P.}~\bsnm{van~der Aalst}},
\bbtitle{Process mining: data science in action},
\bpublisher{Springer Science \& Business Media},
\byear{2016}.
\end{bbook}
\endbibitem

\bibitem{Sztyler2015}
\begin{bchapter}
\bauthor{\binits{T.}~\bsnm{Sztyler}},
\bauthor{\binits{J.}~\bsnm{V{\"o}lker}},
\bauthor{\binits{J.}~\bsnm{Carmona}},
\bauthor{\binits{O.}~\bsnm{Meier}} and
\bauthor{\binits{H.}~\bsnm{Stuckenschmidt}},
\bctitle{Discovery of personal processes from labeled sensor data -- an
  application of process mining to personalized health care},
in: \bbtitle{Proceedings of the International Workshop on Algorithms \&
  Theories for the Analysis of Event Data},
\binstitute{CEUR-ws.org},
\byear{2015},
pp.~\bfpage{22}--\blpage{23}.
\end{bchapter}
\endbibitem

\bibitem{Tax2016a}
\begin{barticle}
\bauthor{\binits{N.}~\bsnm{Tax}},
\bauthor{\binits{N.}~\bsnm{Sidorova}},
\bauthor{\binits{R.}~\bsnm{Haakma}} and
\bauthor{\binits{W.M.P.}~\bsnm{van~der Aalst}},
\batitle{Log-based evaluation of label splits for process models},
\bjtitle{Procedia Computer Science}
\bvolume{96}
(\byear{2016}),
\bfpage{63}--\blpage{72}.
\end{barticle}
\endbibitem

\bibitem{Tax2016b}
\begin{bchapter}
\bauthor{\binits{N.}~\bsnm{Tax}},
\bauthor{\binits{N.}~\bsnm{Sidorova}},
\bauthor{\binits{R.}~\bsnm{Haakma}} and
\bauthor{\binits{W.M.P.}~\bsnm{van~der Aalst}},
\bctitle{Event abstraction for process mining using supervised learning
  techniques},
in: \bbtitle{Proceedings of the SAI Conference on Intelligent Systems},
\bpublisher{Springer},
\byear{2016},
pp.~\bfpage{161}--\blpage{170}.
\end{bchapter}
\endbibitem

\bibitem{Leotta2015}
\begin{bchapter}
\bauthor{\binits{F.}~\bsnm{Leotta}},
\bauthor{\binits{M.}~\bsnm{Mecella}} and
\bauthor{\binits{J.}~\bsnm{Mendling}},
\bctitle{Applying process mining to smart spaces: Perspectives and research
  challenges},
in: \bbtitle{Enterprise, Business-Process and Information Systems Modeling},
\binstitute{Springer},
\byear{2015},
pp.~\bfpage{298}--\blpage{304}.
\end{bchapter}
\endbibitem

\bibitem{Dimaggio2016}
\begin{bchapter}
\bauthor{\binits{M.}~\bsnm{Dimaggio}},
\bauthor{\binits{F.}~\bsnm{Leotta}},
\bauthor{\binits{M.}~\bsnm{Mecella}} and
\bauthor{\binits{D.}~\bsnm{Sora}},
\bctitle{Process-Based Habit Mining: Experiments and Techniques},
in: \bbtitle{Proceedings of the IEEE International Conference on Ubiquitous
  Intelligence \& Computing},
\binstitute{IEEE},
\byear{2016},
pp.~\bfpage{145}--\blpage{152}.
\end{bchapter}
\endbibitem

\bibitem{Sztyler2016}
\begin{bchapter}
\bauthor{\binits{T.}~\bsnm{Sztyler}},
\bauthor{\binits{J.}~\bsnm{Carmona}},
\bauthor{\binits{J.}~\bsnm{V{\"o}lker}} and
\bauthor{\binits{H.}~\bsnm{Stuckenschmidt}},
\bctitle{Self-tracking reloaded: applying process mining to personalized health
  care from labeled sensor data},
in: \bbtitle{Transactions on Petri Nets and Other Models of Concurrency XI},
\bpublisher{Springer},
\byear{2016},
pp.~\bfpage{160}--\blpage{180}.
\end{bchapter}
\endbibitem

\bibitem{Tax2018}
\begin{bchapter}
\bauthor{\binits{N.}~\bsnm{Tax}},
\bauthor{\binits{N.}~\bsnm{Sidorova}},
\bauthor{\binits{R.}~\bsnm{Haakma}} and
\bauthor{\binits{W.M.P.}~\bsnm{van~der Aalst}},
\bctitle{Mining Process Model Descriptions of Daily Life through Event
  Abstraction},
in: \bbtitle{Intelligent Systems and Applications},
\bpublisher{Springer},
\byear{2018},
\bcomment{Chap. To Appear}.
\end{bchapter}
\endbibitem

\bibitem{Huynh2008}
\begin{bchapter}
\bauthor{\binits{T.}~\bsnm{Huynh}},
\bauthor{\binits{M.}~\bsnm{Fritz}} and
\bauthor{\binits{B.}~\bsnm{Schiele}},
\bctitle{Discovery of activity patterns using topic models},
in: \bbtitle{Proceedings of the 10th international conference on Ubiquitous
  computing},
\binstitute{ACM},
\byear{2008},
pp.~\bfpage{10}--\blpage{19}.
\end{bchapter}
\endbibitem

\bibitem{Galushka2006}
\begin{bchapter}
\bauthor{\binits{M.}~\bsnm{Galushka}},
\bauthor{\binits{D.}~\bsnm{Patterson}} and
\bauthor{\binits{N.}~\bsnm{Rooney}},
\bctitle{Temporal data mining for smart homes},
in: \bbtitle{Designing Smart Homes},
\bpublisher{Springer},
\byear{2006},
pp.~\bfpage{85}--\blpage{108}.
\end{bchapter}
\endbibitem

\bibitem{Ogale2007}
\begin{bchapter}
\bauthor{\binits{A.}~\bsnm{Ogale}},
\bauthor{\binits{A.}~\bsnm{Karapurkar}} and
\bauthor{\binits{Y.}~\bsnm{Aloimonos}},
\bctitle{View-invariant modeling and recognition of human actions using
  grammars},
\bpublisher{Springer},
\byear{2007},
pp.~\bfpage{115}--\blpage{126}.
\end{bchapter}
\endbibitem

\bibitem{Peterson1981}
\begin{bbook}
\bauthor{\binits{J.L.}~\bsnm{Peterson}},
\bbtitle{Petri net theory and the modeling of systems},
\bpublisher{Prentice Hall},
\byear{1981}.
\end{bbook}
\endbibitem

\bibitem{Nazerfard2011}
\begin{botherref}
\oauthor{\binits{E.}~\bsnm{Nazerfard}},
\oauthor{\binits{P.}~\bsnm{Rashidi}} and
\oauthor{\binits{D.}~\bsnm{Cook}},
Using association rule mining to discover temporal relations of daily
  activities,
\textit{Toward Useful Services for Elderly and People with Disabilities}
(2011),
49--56.
\end{botherref}
\endbibitem

\bibitem{Jakkula2007}
\begin{bchapter}
\bauthor{\binits{V.R.}~\bsnm{Jakkula}} and
\bauthor{\binits{D.J.}~\bsnm{Cook}},
\bctitle{Using temporal relations in smart environment data for activity
  prediction},
in: \bbtitle{Proceedings of the ICML Workshop on the Induction of Process
  Models},
\byear{2007}.
\end{bchapter}
\endbibitem

\bibitem{Aalst2003}
\begin{barticle}
\bauthor{\binits{W.M.P.}~\bsnm{van Der~Aalst}},
\bauthor{\binits{A.H.M.}~\bsnm{Ter~Hofstede}},
\bauthor{\binits{B.}~\bsnm{Kiepuszewski}} and
\bauthor{\binits{A.P.}~\bsnm{Barros}},
\batitle{Workflow patterns},
\bjtitle{Distributed and parallel databases}
\bvolume{14}(\bissue{1})
(\byear{2003}),
\bfpage{5}--\blpage{51}.
\end{barticle}
\endbibitem

\bibitem{tax2016d}
\begin{bchapter}
\bauthor{\binits{N.}~\bsnm{Tax}},
\bauthor{\binits{E.}~\bsnm{Alasgarov}},
\bauthor{\binits{N.}~\bsnm{Sidorova}} and
\bauthor{\binits{R.}~\bsnm{Haakma}},
\bctitle{On Generation of Time-based Label Refinements},
in: \bbtitle{Proceedings of the 25th International Workshop on Concurrency,
  Specification and Programming},
\byear{2016}.
\end{bchapter}
\endbibitem

\bibitem{Murata1989}
\begin{barticle}
\bauthor{\binits{T.}~\bsnm{Murata}},
\batitle{Petri nets: Properties, analysis and applications},
\bjtitle{Proceedings of the IEEE}
\bvolume{77}(\bissue{4})
(\byear{1989}),
\bfpage{541}--\blpage{580}.
\end{barticle}
\endbibitem

\bibitem{Werf2008}
\begin{barticle}
\bauthor{\binits{J.M.E.M.}~\bsnm{van~der Werf}},
\bauthor{\binits{B.F.}~\bsnm{van Dongen}},
\bauthor{\binits{C.A.J.}~\bsnm{Hurkens}} and
\bauthor{\binits{A.}~\bsnm{Serebrenik}},
\batitle{Process discovery using integer linear programming},
\bjtitle{Fundamenta Informaticae}
\bvolume{94}(\bissue{3--4})
(\byear{2009}),
\bfpage{387}--\blpage{412}.
\end{barticle}
\endbibitem

\bibitem{Goedertier2009}
\begin{barticle}
\bauthor{\binits{S.}~\bsnm{Goedertier}},
\bauthor{\binits{D.}~\bsnm{Martens}},
\bauthor{\binits{J.}~\bsnm{Vanthienen}} and
\bauthor{\binits{B.}~\bsnm{Baesens}},
\batitle{Robust process discovery with artificial negative events},
\bjtitle{Journal of Machine Learning Research}
\bvolume{10}(\bissue{Jun})
(\byear{2009}),
\bfpage{1305}--\blpage{1340}.
\end{barticle}
\endbibitem

\bibitem{Liesaputra2015}
\begin{bchapter}
\bauthor{\binits{V.}~\bsnm{Liesaputra}},
\bauthor{\binits{S.}~\bsnm{Yongchareon}} and
\bauthor{\binits{S.}~\bsnm{Chaisiri}},
\bctitle{Efficient process model discovery using maximal pattern mining},
in: \bbtitle{Proceedings of the International Conference on Business Process
  Management},
\binstitute{Springer},
\byear{2015},
pp.~\bfpage{441}--\blpage{456}.
\end{bchapter}
\endbibitem

\bibitem{Weijters2011}
\begin{bchapter}
\bauthor{\binits{A.}~\bsnm{Weijters}} and
\bauthor{\binits{J.}~\bsnm{Ribeiro}},
\bctitle{Flexible heuristics miner ({FHM})},
in: \bbtitle{Proceedings of the IEEE Symposium on Computational Intelligence
  and Data Mining (CIDM)},
\binstitute{IEEE},
\byear{2011},
pp.~\bfpage{310}--\blpage{317}.
\end{bchapter}
\endbibitem

\bibitem{Augusto2016}
\begin{bchapter}
\bauthor{\binits{A.}~\bsnm{Augusto}},
\bauthor{\binits{R.}~\bsnm{Conforti}},
\bauthor{\binits{M.}~\bsnm{Dumas}},
\bauthor{\binits{M.}~\bsnm{La~Rosa}} and
\bauthor{\binits{G.}~\bsnm{Bruno}},
\bctitle{Automated discovery of structured process models: Discover structured
  vs. discover and structure},
in: \bbtitle{Proceedings of the International Conference on Conceptual
  Modeling},
\binstitute{Springer},
\byear{2016},
pp.~\bfpage{313}--\blpage{329}.
\end{bchapter}
\endbibitem

\bibitem{Dempster1977}
\begin{botherref}
\oauthor{\binits{A.P.}~\bsnm{Dempster}},
\oauthor{\binits{N.M.}~\bsnm{Laird}} and
\oauthor{\binits{D.B.}~\bsnm{Rubin}},
Maximum likelihood from incomplete data via the {EM} algorithm,
\textit{Journal of the Royal Statistical Society. Series B.}
(1977),
1--38.
\end{botherref}
\endbibitem

\bibitem{Kasteren2008}
\begin{bchapter}
\bauthor{\binits{T.}~\bsnm{van Kasteren}},
\bauthor{\binits{A.}~\bsnm{Noulas}},
\bauthor{\binits{G.}~\bsnm{Englebienne}} and
\bauthor{\binits{B.}~\bsnm{Kr{\"o}se}},
\bctitle{Accurate activity recognition in a home setting},
in: \bbtitle{Proceedings of the 10th International Conference on Ubiquitous
  Computing},
\binstitute{ACM},
\byear{2008},
pp.~\bfpage{1}--\blpage{9}.
\end{bchapter}
\endbibitem

\bibitem{Mardia2009}
\begin{bbook}
\bauthor{\binits{K.V.}~\bsnm{Mardia}} and
\bauthor{\binits{P.E.}~\bsnm{Jupp}},
\bbtitle{Directional statistics},
Vol.~\bseriesno{494},
\bpublisher{John Wiley \& Sons},
\byear{2009}.
\end{bbook}
\endbibitem

\bibitem{Rao1976}
\begin{botherref}
\oauthor{\binits{J.}~\bsnm{Rao}},
Some tests based on arc-lengths for the circle,
\textit{Sankhy{\=a}: The Indian Journal of Statistics, Series B}
(1976),
329--338.
\end{botherref}
\endbibitem

\bibitem{Hartigan1985}
\begin{botherref}
\oauthor{\binits{J.A.}~\bsnm{Hartigan}} and
\oauthor{\binits{P.M.}~\bsnm{Hartigan}},
The dip test of unimodality,
\textit{The Annals of Statistics}
(1985),
70--84.
\end{botherref}
\endbibitem

\bibitem{Schwarz1978}
\begin{barticle}
\bauthor{\binits{G.}~\bsnm{Schwarz}},
\batitle{Estimating the dimension of a model},
\bjtitle{The Annals of Statistics}
\bvolume{6}(\bissue{2})
(\byear{1978}),
\bfpage{461}--\blpage{464}.
\end{barticle}
\endbibitem

\bibitem{Kass1995}
\begin{barticle}
\bauthor{\binits{R.E.}~\bsnm{Kass}} and
\bauthor{\binits{A.E.}~\bsnm{Raftery}},
\batitle{Bayes factors},
\bjtitle{Journal of the American Statistical Association}
\bvolume{90}(\bissue{430})
(\byear{1995}),
\bfpage{773}--\blpage{795}.
\end{barticle}
\endbibitem

\bibitem{Cohen2015}
\begin{bchapter}
\bauthor{\binits{T.}~\bsnm{Cohen}} and
\bauthor{\binits{M.}~\bsnm{Welling}},
\bctitle{Harmonic exponential families on manifolds},
in: \bbtitle{Proceedings of The 32nd International Conference on Machine
  Learning},
\binstitute{JMLR W{\&}CP},
\byear{2015},
pp.~\bfpage{1757}--\blpage{1765}.
\end{bchapter}
\endbibitem

\bibitem{Banerjee2005}
\begin{barticle}
\bauthor{\binits{A.}~\bsnm{Banerjee}},
\bauthor{\binits{I.S.}~\bsnm{Dhillon}},
\bauthor{\binits{J.}~\bsnm{Ghosh}} and
\bauthor{\binits{S.}~\bsnm{Sra}},
\batitle{Clustering on the unit hypersphere using von Mises-Fisher
  distributions},
\bjtitle{Journal of Machine Learning Research}
\bvolume{6}(\bissue{Sep})
(\byear{2005}),
\bfpage{1345}--\blpage{1382}.
\end{barticle}
\endbibitem

\bibitem{Hornik2014}
\begin{barticle}
\bauthor{\binits{K.}~\bsnm{Hornik}} and
\bauthor{\binits{B.}~\bsnm{Gr{\"u}n}},
\batitle{movMF: an R package for fitting mixtures of von Mises-Fisher
  distributions},
\bjtitle{Journal of Statistical Software}
\bvolume{58}(\bissue{10})
(\byear{2014}),
\bfpage{1}--\blpage{31}.
\end{barticle}
\endbibitem

\bibitem{Watson1962}
\begin{barticle}
\bauthor{\binits{G.S.}~\bsnm{Watson}},
\batitle{Goodness-of-fit tests on a circle. II},
\bjtitle{Biometrika}
\bvolume{49}(\bissue{1/2})
(\byear{1962}),
\bfpage{57}--\blpage{63}.
\end{barticle}
\endbibitem

\bibitem{Leemans2013}
\begin{bchapter}
\bauthor{\binits{S.J.J.}~\bsnm{Leemans}},
\bauthor{\binits{D.}~\bsnm{Fahland}} and
\bauthor{\binits{W.M.P.}~\bsnm{van~der Aalst}},
\bctitle{Discovering block-structured process models from event logs containing
  infrequent behaviour},
in: \bbtitle{Proceedings of the International Conference on Business Process
  Management},
\binstitute{Springer},
\byear{2013},
pp.~\bfpage{66}--\blpage{78}.
\end{bchapter}
\endbibitem

\bibitem{Rozinat2008}
\begin{barticle}
\bauthor{\binits{A.}~\bsnm{Rozinat}} and
\bauthor{\binits{W.M.P.}~\bsnm{van~der Aalst}},
\batitle{Conformance checking of processes based on monitoring real behavior},
\bjtitle{Information Systems}
\bvolume{33}(\bissue{1})
(\byear{2008}),
\bfpage{64}--\blpage{95}.
\end{barticle}
\endbibitem

\bibitem{Munoz2010}
\begin{bchapter}
\bauthor{\binits{J.}~\bsnm{Munoz-Gama}} and
\bauthor{\binits{J.}~\bsnm{Carmona}},
\bctitle{A fresh look at precision in process conformance},
in: \bbtitle{Proceedings of the International Conference on Business Process
  Management},
\binstitute{Springer},
\byear{2010},
pp.~\bfpage{211}--\blpage{226}.
\end{bchapter}
\endbibitem

\bibitem{Tapia2004}
\begin{bchapter}
\bauthor{\binits{E.M.}~\bsnm{Tapia}},
\bauthor{\binits{S.S.}~\bsnm{Intille}} and
\bauthor{\binits{K.}~\bsnm{Larson}},
\bctitle{Activity recognition in the home using simple and ubiquitous sensors},
in: \bbtitle{Proceedings of the International Conference on Pervasive
  Computing},
\binstitute{Springer},
\byear{2004},
pp.~\bfpage{158}--\blpage{175}.
\end{bchapter}
\endbibitem

\bibitem{Dongen2005}
\begin{botherref}
\oauthor{\binits{B.F.}~\bsnm{van Dongen}},
\oauthor{\binits{A.J.M.M.}~\bsnm{Weijters}} and
\oauthor{\binits{W.M.P.}~\bsnm{van~der Aalst}},
The {P}ro{M} Framework: A New Era in Process Mining Tool Support,
\textit{Applications and Theory of Petri Nets}
(2005),
444.
\end{botherref}
\endbibitem

\bibitem{Herbst2004}
\begin{barticle}
\bauthor{\binits{J.}~\bsnm{Herbst}} and
\bauthor{\binits{D.}~\bsnm{Karagiannis}},
\batitle{Workflow mining with InWoLvE},
\bjtitle{Computers in Industry}
\bvolume{53}(\bissue{3})
(\byear{2004}),
\bfpage{245}--\blpage{264}.
\end{barticle}
\endbibitem

\bibitem{Buijs2012}
\begin{bchapter}
\bauthor{\binits{J.C.A.M.}~\bsnm{Buijs}},
\bauthor{\binits{B.F.}~\bsnm{van Dongen}} and
\bauthor{\binits{W.M.P.}~\bsnm{van~der Aalst}},
\bctitle{On the role of fitness, precision, generalization and simplicity in
  process discovery},
in: \bbtitle{OTM Confederated International Conferences "On the Move to
  Meaningful Internet Systems"},
\binstitute{Springer},
\byear{2012},
pp.~\bfpage{305}--\blpage{322}.
\end{bchapter}
\endbibitem

\bibitem{Li2007}
\begin{bchapter}
\bauthor{\binits{J.}~\bsnm{Li}},
\bauthor{\binits{D.}~\bsnm{Liu}} and
\bauthor{\binits{B.}~\bsnm{Yang}},
\bctitle{Process mining: Extending $\alpha$-algorithm to mine duplicate tasks
  in process logs},
in: \bbtitle{Advances in Web and Network Technologies, and Information
  Management},
\bpublisher{Springer},
\byear{2007},
pp.~\bfpage{396}--\blpage{407}.
\end{bchapter}
\endbibitem

\bibitem{Gu2008}
\begin{bchapter}
\bauthor{\binits{C.-Q.}~\bsnm{Gu}},
\bauthor{\binits{H.-Y.}~\bsnm{Chang}} and
\bauthor{\binits{Y.}~\bsnm{Yi}},
\bctitle{Workflow mining: Extending the $\alpha$-algorithm to mine duplicate
  tasks},
in: \bbtitle{Proceedings of the International Conference on Machine Learning
  and Cybernetics},
Vol.~\bseriesno{1},
\binstitute{IEEE},
\byear{2008},
pp.~\bfpage{361}--\blpage{368}.
\end{bchapter}
\endbibitem

\bibitem{Folino2009}
\begin{bchapter}
\bauthor{\binits{F.}~\bsnm{Folino}},
\bauthor{\binits{G.}~\bsnm{Greco}},
\bauthor{\binits{A.}~\bsnm{Guzzo}} and
\bauthor{\binits{L.}~\bsnm{Pontieri}},
\bctitle{Discovering expressive process models from noised log data},
in: \bbtitle{Proceedings of the International Database Engineering \&
  Applications Symposium},
\binstitute{ACM},
\byear{2009},
pp.~\bfpage{162}--\blpage{172}.
\end{bchapter}
\endbibitem

\bibitem{Vazquez-Barreiros2015}
\begin{bchapter}
\bauthor{\binits{B.}~\bsnm{V\'{a}zquez-Barreiros}},
\bauthor{\binits{M.}~\bsnm{Mucientes}} and
\bauthor{\binits{M.}~\bsnm{Lama}},
\bctitle{Mining duplicate tasks from discovered processes},
in: \bbtitle{Proceedings of the International Workshop on Algorithms \&
  Theories for the Analysis of Event Data},
\binstitute{CEUR-ws.org},
\byear{2015},
pp.~\bfpage{78}--\blpage{82}.
\end{bchapter}
\endbibitem

\bibitem{Lu2016}
\begin{bchapter}
\bauthor{\binits{X.}~\bsnm{Lu}},
\bauthor{\binits{D.}~\bsnm{Fahland}},
\bauthor{\binits{F.J.H.M.}~\bsnm{van~den Biggelaar}} and
\bauthor{\binits{W.M.P.}~\bsnm{van~der Aalst}},
\bctitle{Handling duplicated tasks in process discovery by refining event
  labels},
in: \bbtitle{Proceedings of the International Conference on Business Process
  Management},
\binstitute{Springer},
\byear{2016},
pp.~\bfpage{90}--\blpage{107}.
\end{bchapter}
\endbibitem

\bibitem{DeLeoni2013}
\begin{bchapter}
\bauthor{\binits{M.}~\bsnm{de~Leoni}} and
\bauthor{\binits{W.M.P.}~\bsnm{van~der Aalst}},
\bctitle{Data-aware process mining: discovering decisions in processes using
  alignments},
in: \bbtitle{Proceedings of the 28th Annual ACM Symposium on Applied
  Computing},
\binstitute{ACM},
\byear{2013},
pp.~\bfpage{1454}--\blpage{1461}.
\end{bchapter}
\endbibitem

\bibitem{Allen1983}
\begin{barticle}
\bauthor{\binits{J.F.}~\bsnm{Allen}},
\batitle{Maintaining knowledge about temporal intervals},
\bjtitle{Communications of the ACM}
\bvolume{26}(\bissue{11})
(\byear{1983}),
\bfpage{832}--\blpage{843}.
\end{barticle}
\endbibitem

\bibitem{Gunther2007}
\begin{bchapter}
\bauthor{\binits{C.}~\bsnm{G{\"u}nther}} and
\bauthor{\binits{W.M.P.}~\bsnm{van~der Aalst}},
\bctitle{Fuzzy mining--adaptive process simplification based on
  multi-perspective metrics},
in: \bbtitle{Proceedings of the International Conference on Business Process
  Management},
\bpublisher{Springer},
\byear{2007},
pp.~\bfpage{328}--\blpage{343}.
\end{bchapter}
\endbibitem

\bibitem{Tax2016c}
\begin{barticle}
\bauthor{\binits{N.}~\bsnm{Tax}},
\bauthor{\binits{N.}~\bsnm{Sidorova}},
\bauthor{\binits{R.}~\bsnm{Haakma}} and
\bauthor{\binits{W.M.P.}~\bsnm{van~der Aalst}},
\batitle{Mining local process models},
\bjtitle{Journal of Innovation in Digital Ecosystems}
\bvolume{3}(\bissue{2})
(\byear{2016}),
\bfpage{183}--\blpage{196}.
\end{barticle}
\endbibitem

\bibitem{Reisig1998}
\begin{bbook}
\bauthor{\binits{W.}~\bsnm{Reisig}} and
\bauthor{\binits{G.}~\bsnm{Rozenberg}},
\bbtitle{Lectures on {P}etri nets I: basic models: advances in {P}etri nets},
Vol.~\bseriesno{1491},
\bpublisher{Springer Science \& Business Media},
\byear{1998}.
\end{bbook}
\endbibitem

\bibitem{rcoreteam2013}
\begin{bbook}
\bauthor{\bsnm{{R Core Team}}},
\bbtitle{R: a language and environment for statistical computing},
\binstitute{R Foundation for Statistical Computing},
\blocation{Vienna, Austria}, \byear{2013},
\bcomment{{ISBN} 3-900051-07-0}.
\url{http://www.R-project.org/}.
\end{bbook}
\endbibitem

\bibitem{Benaglia2009}
\begin{barticle}
\bauthor{\binits{T.}~\bsnm{Benaglia}},
\bauthor{\binits{D.}~\bsnm{Chauveau}},
\bauthor{\binits{D.}~\bsnm{Hunter}} and
\bauthor{\binits{D.}~\bsnm{Young}},
\batitle{Mixtools: An R package for analyzing finite mixture models},
\bjtitle{Journal of Statistical Software}
\bvolume{32}(\bissue{6})
(\byear{2009}),
\bfpage{1}--\blpage{29}.
\end{barticle}
\endbibitem

\bibitem{Liu2012}
\begin{bchapter}
\bauthor{\binits{K.}~\bsnm{Liu}},
\bauthor{\binits{W.K.}~\bsnm{Cheung}} and
\bauthor{\binits{J.}~\bsnm{Liu}},
\bctitle{Extracting Behavioral Motifs for Characterizing Human Daily Activities
  in Smart Environments},
in: \bbtitle{Proceedings of the ACM SIGKDD Workshop on Health Informatics},
\byear{2012},
pp.~\bfpage{1}--\blpage{8}.
\end{bchapter}
\endbibitem

\end{thebibliography}

% or include bibliography directly:
%\begin{thebibliography}{0}
%\bibitem{r1} F. Author, Information about cited object.
%
%\bibitem{r2} S. Author and T. Author, Information about cited object.
%\end{thebibliography}

\end{document}